\documentclass[10pt,journal,compsoc]{IEEEtran}
\ifCLASSOPTIONcompsoc
  \usepackage[nocompress]{cite}
\else
  \usepackage{cite}
\fi
\ifCLASSINFOpdf
\else
\fi

\usepackage{epsfig}
\usepackage{graphicx}
\usepackage{amsmath}
\usepackage{amssymb}
\usepackage{enumitem}
\usepackage{adjustbox}
\usepackage{lscape}
\usepackage[table,xcdraw]{xcolor}
\usepackage{booktabs}
\usepackage{array} 
\usepackage{longtable}
\usepackage{colortbl}
\usepackage{multirow}
\usepackage{lscape}
\usepackage{makecell}
\usepackage{tabularx}
\usepackage[utf8]{inputenc}

\usepackage[pagebackref=true,breaklinks=true,letterpaper=true,colorlinks,bookmarks=false,urlcolor=red,citecolor=cyan]{hyperref}
\usepackage{gensymb,graphics,inputenc}
\usepackage[linesnumbered,algo2e,boxed]{algorithm2e}
\graphicspath{{Figure/}}
\usepackage[figuresright]{rotating}
\usepackage{amsmath,amssymb,textcomp,multirow,upgreek}
\usepackage{booktabs}
\usepackage{color,mdwlist}

\usepackage{ragged2e}

\usepackage{review}

\usepackage{graphicx,fontawesome}
\graphicspath{{Figures/}}  
\usepackage{subcaption}

\usepackage{caption}
\usepackage{wrapfig}

\usepackage{enumitem}
\setitemize{noitemsep,topsep=0pt,parsep=0pt,partopsep=0pt}

\usepackage{tikz}
\usetikzlibrary{positioning, arrows.meta}

\usepackage{ifthen}

\definecolor{SciBlue}{RGB}{0,84,159}        
\definecolor{SciPurple}{RGB}{146,39,143}    
\definecolor{SciOrange}{RGB}{255,140,0}     
\definecolor{SciGreen}{RGB}{0,128,55}       
\definecolor{SciGray}{RGB}{89,89,89}     

\newcommand{\timelineEvent}[7]{
    \ifstrequal{#2}{up}{\def\direction{1} \def\anchor{south}}{\def\direction{-1} \def\anchor{north}}
    \draw[-{Stealth[length=2mm, width=1.5mm]}, color=#4, thick] (#1, 0) -- (#1, \direction*#3);
    \node[anchor=\anchor, inner sep=2pt, align=center, font=\small, text=#7] at (#1, \direction*#3) {#5 \\ #6};
}

\begin{document}
%
\title{Multimodal Data Storage and Retrieval for Embodied AI: A Survey}
\author{Yihao Lu \quad Hao Tang
	\IEEEcompsocitemizethanks{
	    \IEEEcompsocthanksitem Yihao Lu is with the School of Economics and Management, South China Normal University, Guangzhou 511436, China.
	    E-mail: 20230734031@m.scnu.edu.cn \protect
        \IEEEcompsocthanksitem Hao Tang is with the State Key Laboratory of Multimedia Information Processing,
School of Computer Science, Peking University, China. Email: haotang@pku.edu.cn \protect
	    \IEEEcompsocthanksitem Corresponding author: Hao Tang. \protect
        }
}

%
%

\markboth{IEEE Transactions on Pattern Analysis and Machine Intelligence}%
{Shell \MakeLowercase{\textit{et al.}}: Bare Demo of IEEEtran.cls for Computer Society Journals}
%



\IEEEtitleabstractindextext{%
\justify
\begin{abstract}
Embodied AI (EAI) agents continuously interact with the physical world, generating vast, heterogeneous multimodal data streams that traditional management systems are ill-equipped to handle. In this survey, we first systematically evaluate five storage architectures (Graph Databases, Multi-Model Databases, Data Lakes, Vector Databases, and Time-Series Databases), focusing on their suitability for addressing EAI's core requirements, including physical grounding, low-latency access, and dynamic scalability. We then analyze five retrieval paradigms (Fusion Strategy-Based Retrieval, Representation Alignment-Based Retrieval, Graph-Structure-Based Retrieval, Generation Model-Based Retrieval, and Efficient Retrieval-Based Optimization), revealing a fundamental tension between achieving long-term semantic coherence and maintaining real-time responsiveness. Based on this comprehensive analysis, we identify key bottlenecks, spanning from the foundational Physical Grounding Gap to systemic challenges in cross-modal integration, dynamic adaptation, and open-world generalization. Finally, we outline a forward-looking research agenda encompassing physics-aware data models, adaptive storage-retrieval co-optimization, and standardized benchmarking, to guide future research toward principled data management solutions for EAI. Our survey is based on a comprehensive review of more than 180 related studies, providing a rigorous roadmap for designing the robust, high-performance data management frameworks essential for the next generation of autonomous embodied systems.
\end{abstract}

}

\maketitle

\IEEEdisplaynontitleabstractindextext

%
\IEEEpeerreviewmaketitle


%
%
%
%

\section{Introduction}
\label{Introduction}

\begin{figure*}[t!]
    \centering
    \resizebox{0.95\textwidth}{!}{
        \begin{minipage}{\textwidth}
            \begin{subfigure}[b]{0.33\textwidth}
                \centering
                \includegraphics[height=13cm]{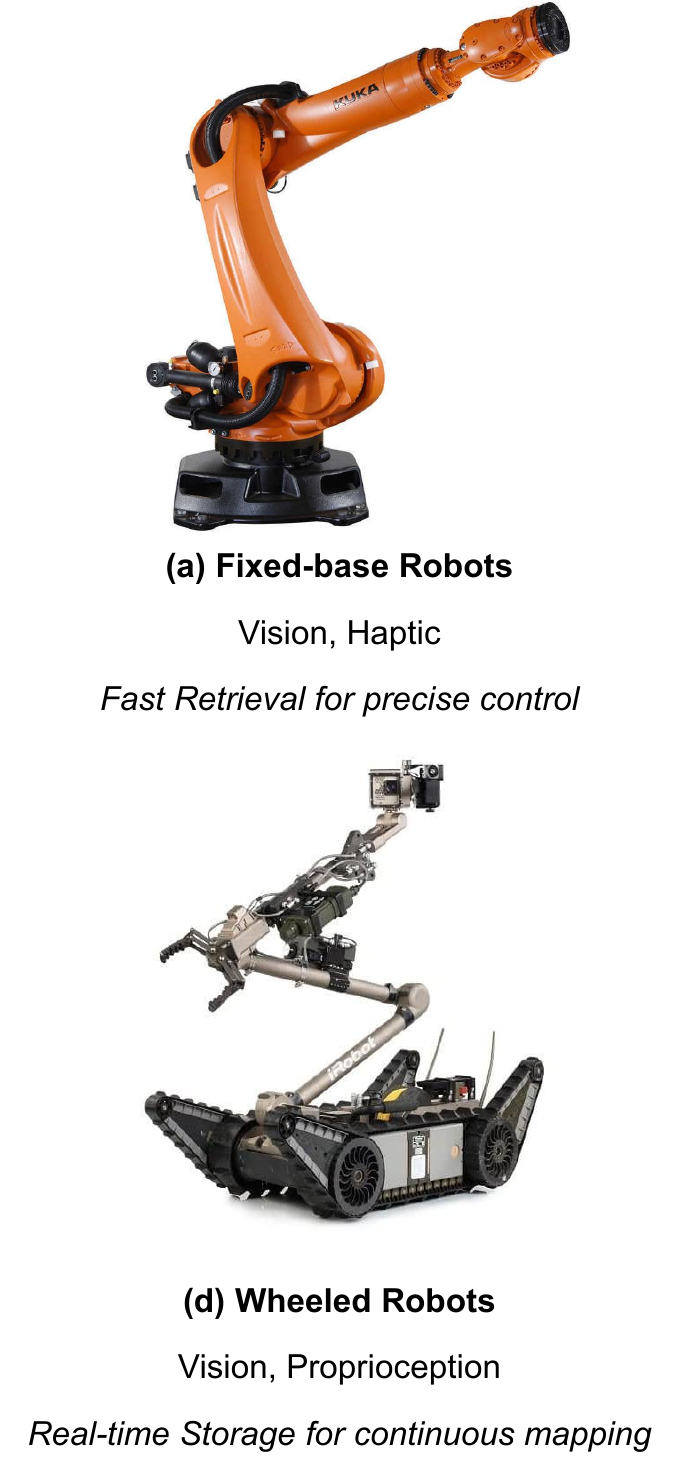}
            \end{subfigure}
            \hfill 
            \begin{subfigure}[b]{0.33\textwidth}
                \centering
                \includegraphics[height=13cm]{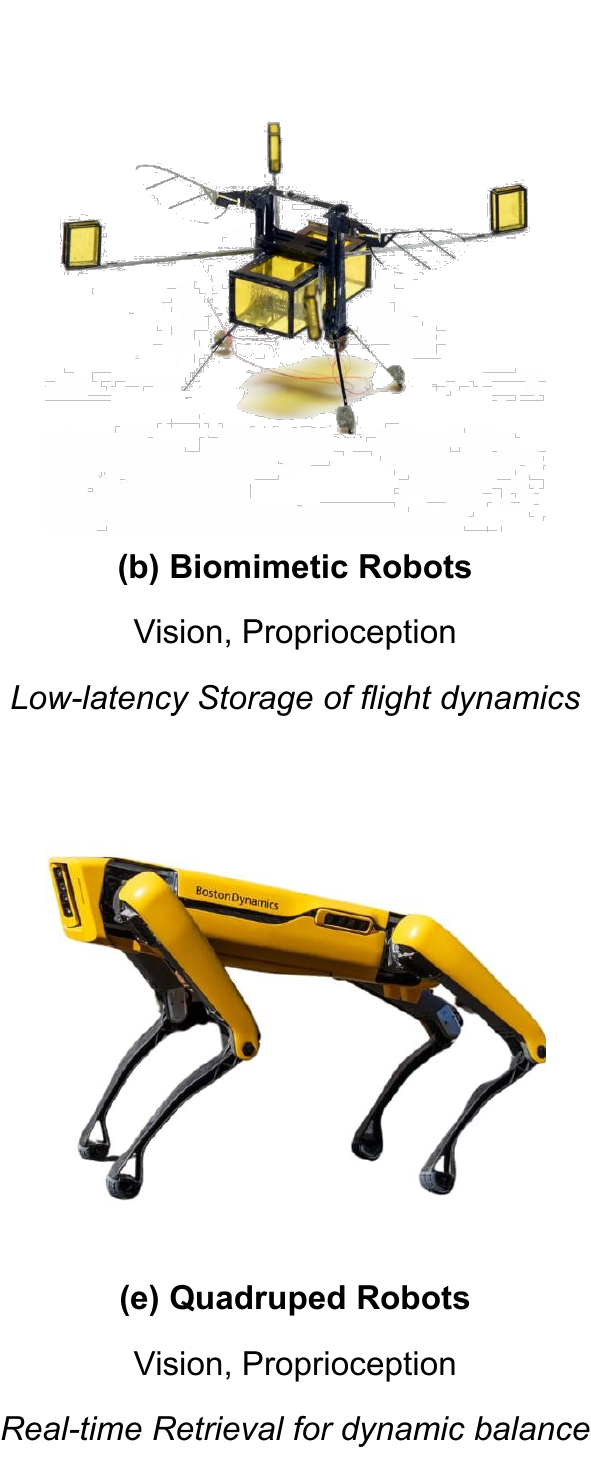}
            \end{subfigure}
            \hfill
            \begin{subfigure}[b]{0.33\textwidth}
                \centering
                \includegraphics[height=13cm]{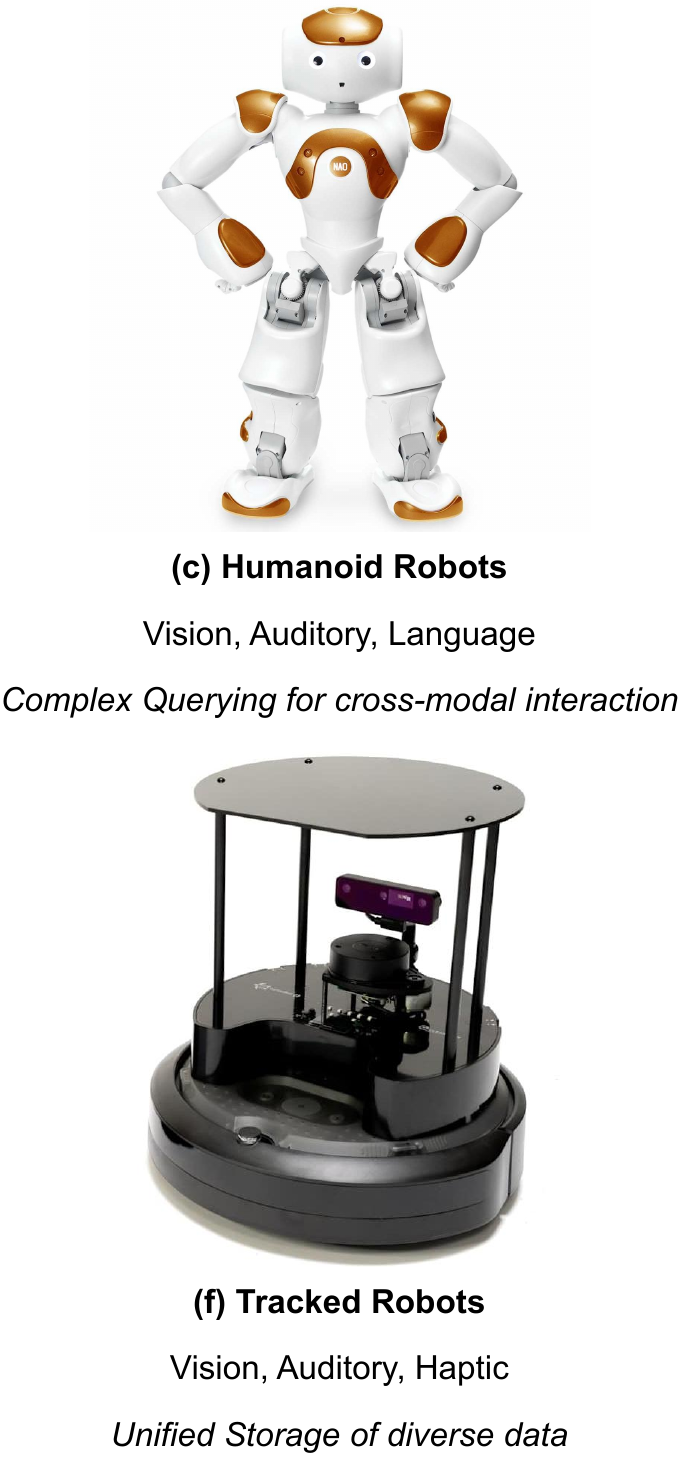}
            \end{subfigure}
        \end{minipage}
    }
    \caption{Illustrative examples of diverse physical morphologies in Embodied AI agents, including (a) Fixed-base Robots, (b) Biomimetic Robots, (c) Humanoid Robots, (d) Wheeled Robots, (e) Quadruped Robots, and (f) Tracked Robots. For each agent, the first line of text under its title indicates the core data modalities it generates, while the second, \textit{italicized} line specifies its primary data management need. The classification of agent morphologies is adapted from~\cite{liu2024aligningcyberspacephysical}.}
    \label{Illustrative_examples_of_diverse_physical_morphologies_in_Embodied_AI}
\end{figure*}

Among AI’s various subfields, embodied intelligence refers to agents that learn by directly interacting with their physical surroundings. Many researchers view embodied AI as essential for realizing artificial general intelligence~\cite{Gupta_2021,roy2021machinelearningroboticschallenges}. Unlike classical AI---which relies on abstract computation and massive datasets---embodied AI emphasizes an agent’s capacity to enact and adapt behaviors in real-world environments~\cite{zheng2024surveyembodiedlearningobjectcentric}. Beyond dialogue systems like ChatGPT, a comprehensive vision of AGI includes the ability to control physical agents and interact deeply with both simulated and real environments~\cite{duan2022surveyembodiedaisimulators,xu2024surveyroboticsfoundationmodels,ma2024surveyvisionlanguageactionmodelsembodied}. These entities, termed Embodied AI (EAI) agents, span a wide range of physical morphologies (Fig.~\ref{Illustrative_examples_of_diverse_physical_morphologies_in_Embodied_AI}). They range from static industrial arms and bio-inspired microrobots to complex humanoid and mobile platforms. This diversity is not merely superficial; it fundamentally dictates the nature and complexity of the multimodal data each agent generates and must process to interact with the world.

A model's generalization ability is critically influenced by the size and quality of its training data. Scaling laws show that larger models require more data to enable complex environmental adaptation and robust task generalization~\cite{kaplan2020scalinglawsneurallanguage}. However, Lin et al.~\cite{lin2024datascalinglawsimitation} show that generalization scales as a power law with both environmental variety and object count. This suggests that diversity is often more critical than sheer quantity: once a threshold of examples per environment or object is reached, additional demonstrations yield diminishing returns.

However, EAI presents unique data-centric challenges beyond sheer scale. First, agents must process heterogeneous data streams, including sensory inputs (e.g., vision, touch), motor commands, and environmental feedback, in a continuous loop to enable real-time learning and adaptation~\cite{zheng2024surveyembodiedlearningobjectcentric}. Second, they operate in complex, dynamic environments, where data characteristics can shift unpredictably. This imposes stringent requirements not only on data volume but also on its quality and contextual relevance, particularly in safety-critical applications like autonomous driving. As model scales expand, the effective collection, management, and processing of high-quality data are becoming central bottlenecks that hinder the advancement of the field.

From autonomous vehicles processing terabytes of sensor data in real-time to surgical robots demanding flawless data transmission for remote operations, Embodied Intelligence manifests in a wide array of data-intensive applications. These representative scenarios underscore the critical need for advanced multimodal data storage and retrieval technologies---the focus of this survey.

\subsection{Research Motivation and Goals}

\begin{figure}
    \centering
    \includegraphics[width=1\linewidth]{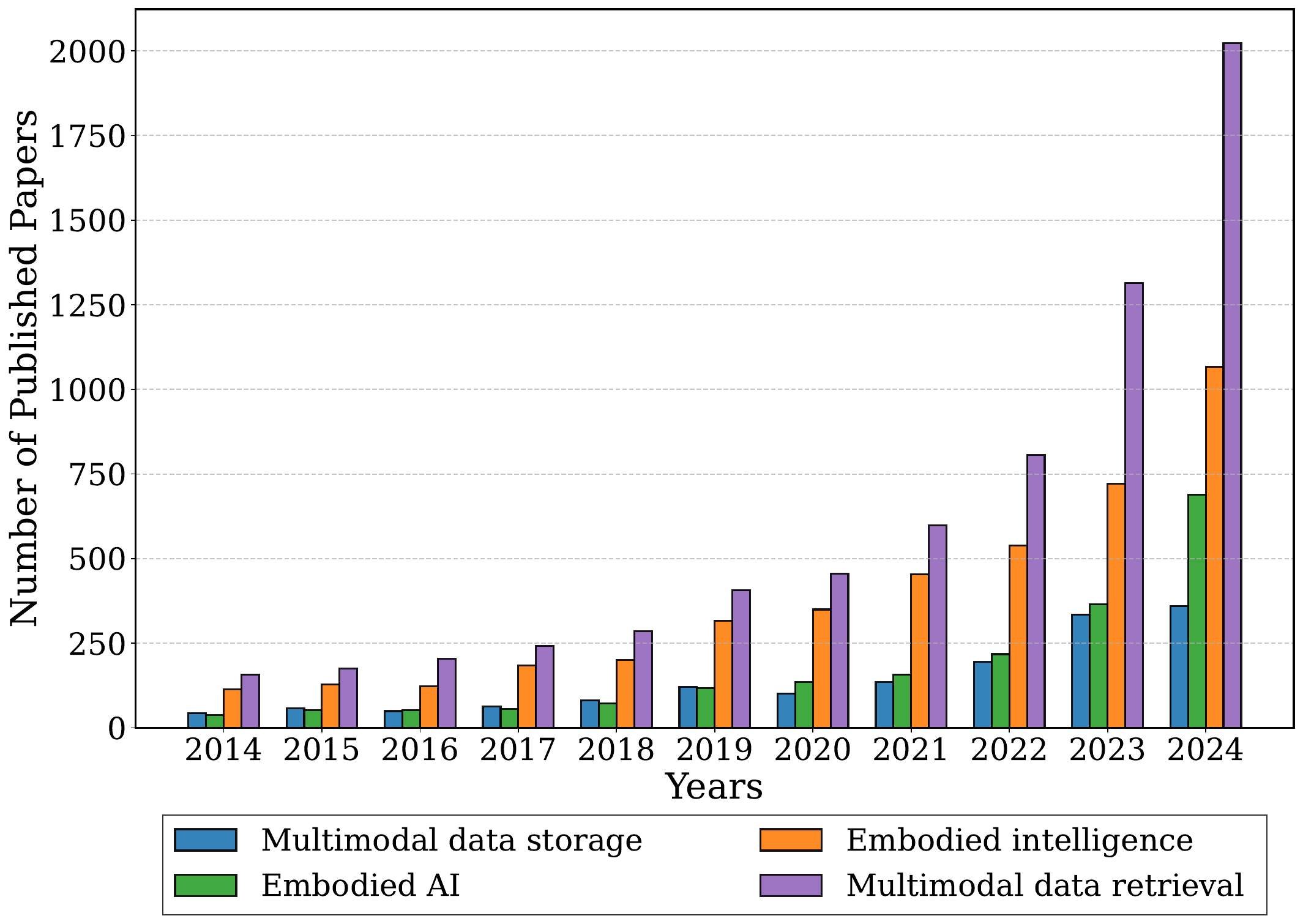}
    \caption{Growth in publications on Web of Science for key research topics: `Multimodal data storage,' `Embodied intelligence,' `Embodied AI,' and `Multimodal data retrieval' (2014-2024).}
    \label{wos_numbers_of_paper}
    \vspace{-0.4cm}
\end{figure}

The rapid growth in publications concerning Embodied AI and multimodal data management underscores the pressing need for this survey, a trend quantified in Fig.~\ref{wos_numbers_of_paper}. The confluence of these burgeoning fields necessitates a systematic review of the technologies that bridge them.

This survey critically examines multimodal data storage and retrieval methods tailored for embodied intelligence, synthesizing recent advancements and highlighting key research opportunities moving forward. This paper makes three main contributions:

\begin{enumerate}[leftmargin=*]
    \item \textbf{Filling the research gap:} Despite the heavy reliance of embodied intelligence on multimodal data processing, comprehensive reviews of storage and retrieval strategies tailored to its interactive and physically grounded nature remain limited.
    \item \textbf{Fostering cross-domain innovation:} We examine how established multimodal data methods can be adapted to the distinct real-time, spatial, and sensory processing challenges encountered in embodied intelligence systems.
    \item \textbf{Identifying future directions:} By analyzing existing bottlenecks, we propose targeted optimization strategies and delineate promising directions for future research.
\end{enumerate}

To provide a structured overview, we present a strategic summary in Table~\ref{Strategic_Positioning_and_Core_Challenges_of_Data_Management_Technologies_for_Embodied_AI}. This table highlights key technology paradigms, their strategic roles in EAI, core capabilities, trade-offs, and future research frontiers. It acts as a roadmap for the in-depth discussions in the subsequent sections.

\begin{table*}[htbp]
  \centering
  \small
  \caption{Strategic Positioning and Core Challenges of Data Management Technologies for Embodied AI}
  \label{Strategic_Positioning_and_Core_Challenges_of_Data_Management_Technologies_for_Embodied_AI}
  \renewcommand{\arraystretch}{1.5}
  \resizebox{\textwidth}{!}{
  \begin{tabular}{
      >{\centering\arraybackslash}p{3.5cm}
      >{\centering\arraybackslash}p{3cm}
      >{\raggedright\arraybackslash}p{4.5cm}
      >{\centering\arraybackslash}p{3.5cm}
      >{\raggedright\arraybackslash}p{4.8cm}
    }
    \toprule
    \multicolumn{1}{c}{\textbf{Technology Paradigm}} & 
    \multicolumn{1}{c}{\textbf{Strategic Role in EAI}} & 
    \multicolumn{1}{c}{\textbf{Core Capability}} & 
    \multicolumn{1}{c}{\textbf{Key Trade-off}} & 
    \multicolumn{1}{c}{\textbf{Future Research Frontier}} \\
    \midrule
    \multicolumn{5}{l}{\textit{Multimodal Data Storage (\S\ref{Multimodal_Data_Storage})}} \\ 
    \midrule

    \textbf{Graph Databases} \par (\S\ref{Graph_Databases}) 
    & \textbf{The Context Modeler} & Models entity relationships and enables causal reasoning. & \textbf{Expressiveness vs. Latency} & Real-time graph updates under continuous data streams. \\

    \textbf{Multi-model Databases} \par (\S\ref{Multi-Model_Databases})
    & \textbf{The Unified Hub} & Consolidates diverse data types under a single interface. & \textbf{Generality vs. Performance} & Context-aware schema evolution for dynamic tasks. \\

    \textbf{Data Lakes} \par (\S\ref{Data_Lakes})
    & \textbf{The Raw Experience Archive} & Stores massive, unstructured sensory data for offline analysis. & \textbf{Retention vs. Accessibility} & Bridging offline analysis with low-latency online decision loops. \\

    \textbf{Vector Databases} \par (\S\ref{Vector_Databases})
    & \textbf{The Semantic Memory} & Enables similarity search over learned multimodal embeddings. & \textbf{Scalability vs. Index Freshness} & Maintaining retrieval stability during dynamic environmental shifts. \\

    \textbf{Time-Series Databases} \par (\S\ref{Time-Series_Databases})
    & \textbf{The Dynamic State Tracker} & Manages high-frequency, time-stamped sensor data. & \textbf{Temporal Focus vs. Holistic View} & Cross-modal fusion of temporal and non-temporal data. \\
    \midrule
    \multicolumn{5}{l}{\textit{Multimodal Data Retrieval (\S\ref{Multimodal_Data_Retrieval})}} \\
    \midrule
    \textbf{Fusion Strategy-Based Retrieval} \par (\S\ref{Fusion_Strategy-Based_Retrieval})
    & \textbf{The Coherence Engine} & Combines features from multiple modalities to enhance perception. & \textbf{Complexity vs. Efficiency} & Lightweight, adaptive fusion mechanisms for edge devices. \\

    \textbf{Representation Alignment-Based Retrieval} \par (\S\ref{Representation_Alignment-Based_Retrieval})
    & \textbf{The  ``Common Language'' Creator} & Maps heterogeneous data into a unified semantic space. & \textbf{Consistency vs. Adaptability} & Online alignment adaptation to combat semantic drift. \\

    \textbf{Graph-Structure-Based Retrieval} \par (\S\ref{Graph-Structure-Based_Retrieval})
    & \textbf{The Relational Reasoner} & Infers multi-step plans by traversing the knowledge graph. & \textbf{Depth vs. Speed} & Efficiently querying large, sparse, and long-tail event graphs. \\

    \textbf{Generation Model-Based Retrieval} \par (\S\ref{Generation_Model-Based_Retrieval})
    & \textbf{The ``What-if'' Simulator} & Fills information gaps and generates contextual responses. & \textbf{Creativity vs. Factual Grounding} & Balancing low-latency response with mitigating ``hallucinations''. \\

    \textbf{Efficient Retrieval-Based Optimization} \par (\S\ref{Efficient_Retrieval-Based_Optimization})
    & \textbf{The Real-time Accelerator} & Ensures fast, approximate retrieval on resource-constrained platforms. & \textbf{Speed vs. Precision} & Dynamic index updating strategies aware of environmental changes. \\
    \bottomrule
  \end{tabular}
  }
\end{table*}

\subsection{Paper Organization}

The remainder of this paper is organized as follows:

\begin{itemize}[leftmargin=*]
    \item \textbf{Sec.~\ref{Embodied_AI_and_multimodal_data}} surveys the fundamentals of embodied intelligence and analyzes the unique characteristics and processing requirements of the data it generates.
    \item \textbf{Sec.~\ref{Multimodal_Data_Storage}} investigates various multimodal data storage technologies, evaluating the suitability of different architectures and data models for EAI applications.
    \item \textbf{Sec.~\ref{Multimodal_Data_Retrieval}} analyzes key paradigms for multimodal data retrieval, with a focus on their effectiveness in addressing requirements for semantic understanding, cross-modal association, and real-time performance.
    \item \textbf{Sec.~\ref{Discussion}} distills the preceding analyses into a discussion of the overarching challenges and open problems in applying existing storage and retrieval technologies to the EAI domain.
    \item \textbf{Sec.~\ref{Conclusion}} concludes the survey with a summary of key findings and outlines promising directions for future research.
\end{itemize}

\section{Embodied AI and Multimodal Data}
\label{Embodied_AI_and_multimodal_data}

\subsection{Theoretical Foundations of Embodied Intelligence}

The conceptual foundations of embodied intelligence trace back to the earliest days of AI research. In his 1950 paper ``Computing Machinery and Intelligence''~\cite{1950Computing}, Turing proposed two complementary paradigms: one emphasizing abstract computation and the other equipping machines with sensors to learn through direct interaction (see Table~\ref{Comparison_between_disembodied_AI_and_embodied_AI} for a comparison). 

\begin{table*}[htbp]
  \centering
  \caption{Comparison between disembodied AI and embodied AI (adapted from~\cite{liu2024aligningcyberspacephysical})}
  \label{Comparison_between_disembodied_AI_and_embodied_AI}
  \resizebox{\textwidth}{!}{
  \begin{tabular}{
      >{\centering\arraybackslash}m{2cm}
      >{\centering\arraybackslash}m{2cm}
      >{\centering\arraybackslash}m{2.5cm}
      >{\centering\arraybackslash}m{4.5cm}
      >{\centering\arraybackslash}m{3.5cm}
      >{\centering\arraybackslash}m{4cm}
    }
    \toprule
    \textbf{Type} & \textbf{Environment} & \textbf{Physical Entities} & \textbf{Description} & \textbf{Primary Data Types} & \textbf{Representative Agents} \\
    \midrule
    Disembodied AI & Cyber Space & No & Cognition and physical entities are disentangled & Vision, Auditory, Language & ChatGPT~\cite{achiam2023gpt}, RoboGPT~\cite{chen2023robogpt} \\
    \midrule
    Embodied AI & Physical Space & Robots, Cars, Other devices & Cognition is integrated into physical entities & Vision, Auditory, Language, \textbf{Haptic} & RT-1~\cite{brohan2022rt}, RT-2~\cite{brohan2023rt}, RT-H~\cite{belkhale2024rt} \\
    \bottomrule
  \end{tabular}
  }
\end{table*}

In ``Intelligence Without Representation'' (1991)~\cite{1991Intelligence}, Rodney Brooks challenged classic AI theory by showing that complex behaviors can emerge from simple body–environment interactions, without rich internal models. His ``Behaviorist Intelligence'' framework has since inspired many autonomous robot designs and spurred research on ``Underlying Intelligence,'' in which layered perceptual–motor responses give rise to higher-level behaviors.

In 1999, Rolf Pfeiffer and Christian Schell extended this view in \textit{Understanding Intelligence}~\cite{richardson2022understanding}, introducing ``Physical Intelligence''. They argue that intelligence arises from an agent’s body structure and its environment, not solely from brain-like computation. Their ``Morphological Computing'' concept shows how leveraging an agent's physical form can simplify control and perception tasks. 

The evolution of EAI from simple reactive controllers to complex, multimodal agents has introduced significant data-management challenges in real-world settings. Section~\ref{Representative_Application_Scenarios_in_EAI} illustrates key application scenarios that drive the need for advanced storage and retrieval solutions. 

\subsection{Representative Application Scenarios in EAI}
\label{Representative_Application_Scenarios_in_EAI}

To ground our survey, we first examine how real-world embodied systems generate, store, and query vast multimodal data streams across several key domains. 

\subsubsection{Industrial Manufacturing}

In industrial manufacturing, embodied intelligence is advancing robots from merely active to genuinely capable of working~\cite{CAICT2024Embodied}. For instance, using frameworks like ChatGPT to control robotic arms, drones, and mobile robots~\cite{vemprala2023chatgptroboticsdesignprinciples} requires the real-time integration of diverse data. Robots in hazardous environments, aimed at improving safety and efficiency~\cite{Zhao2024Binocular}, must fuse visual data for navigation, haptic data for manipulation, and often linguistic data for human collaboration. The core data management challenge lies in synchronously processing these streams for immediate action while archiving interaction logs for long-term process optimization and skill refinement, especially in low-volume, high-variation production environments like Evolvable Assembly Systems~\cite{frei2008embodied}.

\subsubsection{Autonomous Driving}

Autonomous driving exemplifies a demanding multimodal data ecosystem, requiring both ultra-low-latency local perception and massive offline repositories for long-term learning and simulation. Early benchmarks such as KITTI~\cite{geiger2012we} established the field’s evaluation conventions, paving the way for modern large-scale, multi-sensor datasets. These include nuScenes~\cite{caesar2020nuscenes} and the Waymo Open Dataset~\cite{sun2020scalability}, which provide camera, LiDAR, and radar streams that enable robust perception, tracking, and forecasting research. Contemporary work has moved beyond single-vehicle perception to cooperative V2X reconstruction and editing: for example, CRUISE~\cite{xu2025cruise} demonstrates decomposed Gaussian-splat representations to collaboratively reconstruct and edit driving scenes across infrastructure and vehicles. This directly motivates data architectures that support distributed indexing, low-latency cross-agent queries, and fidelity-preserving compression.  Such systems, exemplified by Tesla's Autopilot ``HydraNet''~\cite{Chen2022TheIO}, which ingests high-resolution video streams, demand two complementary data tiers:
1) An ultra-low-latency storage and retrieval layer for safety-critical, real-time perception and control;
2) A massive, scalable repository for offline analysis, simulation, and the training of next-generation driving models.

\subsubsection{Service Robotics}
Service robots further illustrate these needs, with examples including:

\begin{enumerate}[leftmargin=*]
    \item \textbf{Logistics and Warehousing:} Agility Robotics' humanoid robot, Digit, autonomously navigates warehouses, requiring the continuous processing of visual and proprioceptive data to maintain balance and navigate dynamic spaces.
    \item \textbf{Home and Personal Assistance:} In home settings, robots like Mobile ALOHA~\cite{fu2024mobilealohalearningbimanual} and ALOHA2~\cite{aloha2team2024aloha2enhancedlowcost} perform complex chores by fusing visual data with fine-grained haptic feedback for dexterous manipulation.
    \item \textbf{Social and Companion Robots:} To provide emotional companionship, social robots such as LOVOT must interpret subtle, real-time multimodal cues, including visual (e.g., facial expressions) and auditory (e.g., tone of voice) signals, to engage with users effectively~\cite{9791008, klodmann2021introduction}.
\end{enumerate}

Across all these service applications, a common data challenge is the management of long-term, unstructured interaction data, which is crucial for enabling robust, lifelong learning and adaptation in dynamic human environments.

\subsubsection{Other Domains}
In healthcare, the da Vinci surgical robot demands highly reliable, low-latency access to high-definition stereo video and haptic signals for remote operations. Moreover, assistive robots in psychotherapy and elder care process highly sensitive multimodal data, raising significant privacy and compliance concerns~\cite{doi:10.1089/tmj.2018.0051}. Recent work also highlights the value of online adaptive fusion in clinical settings. TTTFusion~\cite{xie2025tttfusiontesttimetrainingbasedstrategy} employs test-time training to adapt multimodal medical image fusion and improve robustness at deployment. Forestry and agriculture likewise depend on robust sensor processing, with applications ranging from tree-to-tree locomotion machines~\cite{parker2016robotics} to agricultural robots for weed control~\cite{aastrand2002agricultural, BLASCO2002149} that must navigate and operate in complex natural environments. The primary data challenge across these domains is to guarantee reliability, safety, and precision in managing high-stakes, environment-grounded data.

\subsection{Definition and Types of Multimodal Data}

Multimodal data refers to the integration of distinct data types---such as vision, language, hearing, touch, and even smell-each offering unique, complementary information that can be combined to enhance system understanding. For instance, in a manipulation task, visual data from cameras captures an object's location and appearance, while haptic feedback from tactile sensors reveals its texture and rigidity. Fusing these data streams creates a far more comprehensive environmental model than any single modality could provide alone.

\begin{figure}
    \centering
    \includegraphics[width=1\linewidth]{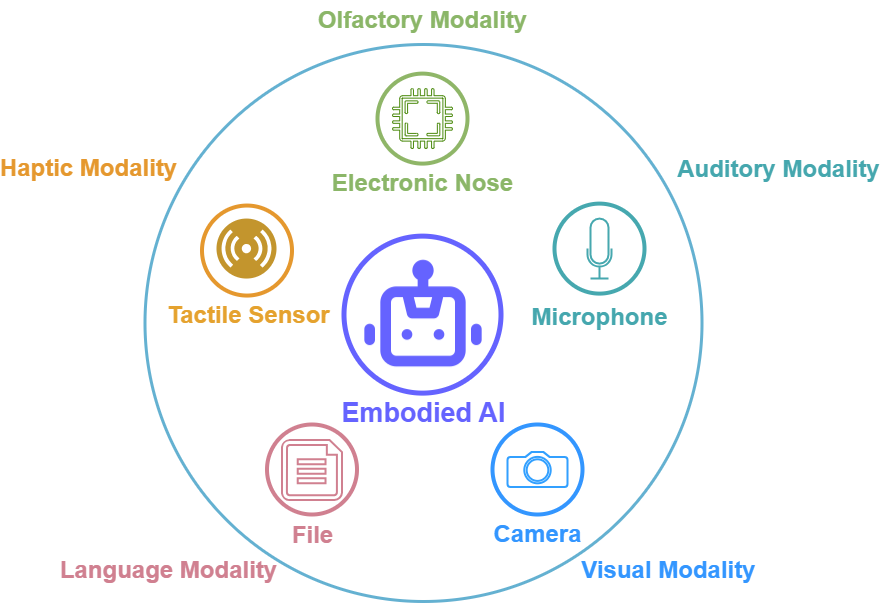}
    \caption{Overview of Sensory Modalities and Data Sources for Embodied AI}
    \label{Overview_of_Sensory_Modalities_and_Data_Sources_for_Embodied_AI}
\end{figure}

Fig.~\ref{Overview_of_Sensory_Modalities_and_Data_Sources_for_Embodied_AI} presents a conceptual overview of these key sensory modalities and their corresponding data sources. A fundamental challenge in EAI data management, which this survey addresses, lies in effectively storing and retrieving these distinct yet interconnected data streams. Below, we detail the primary characteristics and roles of each major modality.

\subsubsection{Visual Modalities}

Visual modalities remain the central perceptual channel for embodied agents. Recent advances in visual representation learning driven by Vision Transformer~\cite{dosovitskiy2020image} and detection transformers~\cite{carion2020end} have reshaped how image and scene features are extracted and reasoned over. Concurrently, vision-language alignment models like CLIP~\cite{radford2021learning} provide semantic embeddings that support cross-modal retrieval and zero-shot generalization. For tasks that require geometric reasoning, PointNet~\cite{qi2017pointnet} and its successors supply efficient encoders for point clouds and 3D geometry that interface naturally with 2D vision–language backbones. More recent 3D vision–language models, including 3D-R1~\cite{huang20253dr1enhancingreasoning3d}, integrate geometric representations with language priors to improve unified scene reasoning. This motivates storage and retrieval designs that combine dense semantic indices with structured, geometry-aware indices and mechanisms for aligning semantic and geometric matches at query time.

\subsubsection{Language Modalities}

Language modalities, which comprise text and speech, are the principal human-agent interface. Modern embodied systems commonly combine robust speech transcription backbones like Whisper~\cite{radford2023robust} and wav2vec 2.0~\cite{baevski2020wav2vec} with large language or embodied multimodal models, including PaLM-E~\cite{10.5555/3618408.3618748} and SayCan~\cite{brohan2023can}. These models ground high-level instructions to actionable skill sequences. This stack of technologies from automatic speech recognition (ASR) to skill grounding reduces the symbol grounding gap and allows downloadable language traces to be stored in semantic indices for retrospective retrieval, instruction replay, and offline behavior cloning.

\subsubsection{Auditory Modalities}

Auditory signals complement vision by providing out-of-sight cues (sirens, colliding sounds, speech) and enable applications in audio-guided navigation and event detection. Large datasets like AudioSet~\cite{gemmeke2017audio} and pretrained audio backbones like PANNs~\cite{kong2020panns} furnish generic audio embeddings that can be indexed for retrieval. In embodied navigation, simulators, notably SoundSpaces~\cite{chen2020soundspaces}, show how audio-visual joint training materially improves localization and goal-directed search. For storage and retrieval design, audio introduces long-duration, high-sample-rate streams that benefit from specialized compression, time-series indexing, and segment-level semantic indexing for efficient cross-modal joins with vision and language records.

\subsubsection{Haptic Modalities}

Haptic sensing provides contact-level, material and force information that is essential for fine manipulation. Vision-based tactile sensors (GelSight)~\cite{yuan2017gelsight} capture high-resolution geometry and slip/force cues; standard manipulation benchmarks such as the YCB Object Set~\cite{calli2015ycb} provide common objects for comparative evaluation. For multimodal data systems, tactile streams are typically high-bandwidth at short time-scales and may be stored as event-triggered traces or aggregated feature summaries (tactile embeddings) to enable efficient retrieval and replay during manipulation learning.

\subsubsection{Olfactory Modalities}

The olfactory modality, while less common, offers unique sensory information. The development of the electronic nose~\cite{1982Analysis} and advanced algorithms~\cite{Imam_2020} has made machine-based olfaction feasible. Projects like Osmo, which use graph neural networks to digitize odors, are pushing the frontier of odor digitization and representation. In EAI, olfaction can provide critical environmental cues, such as detecting hazardous gas leaks in industrial settings or assessing food freshness in automated kitchens.

\subsection{The Role of Multimodal Data in Embodied Intelligence}

\begin{figure*}
    \centering
    \includegraphics[width=1\linewidth]{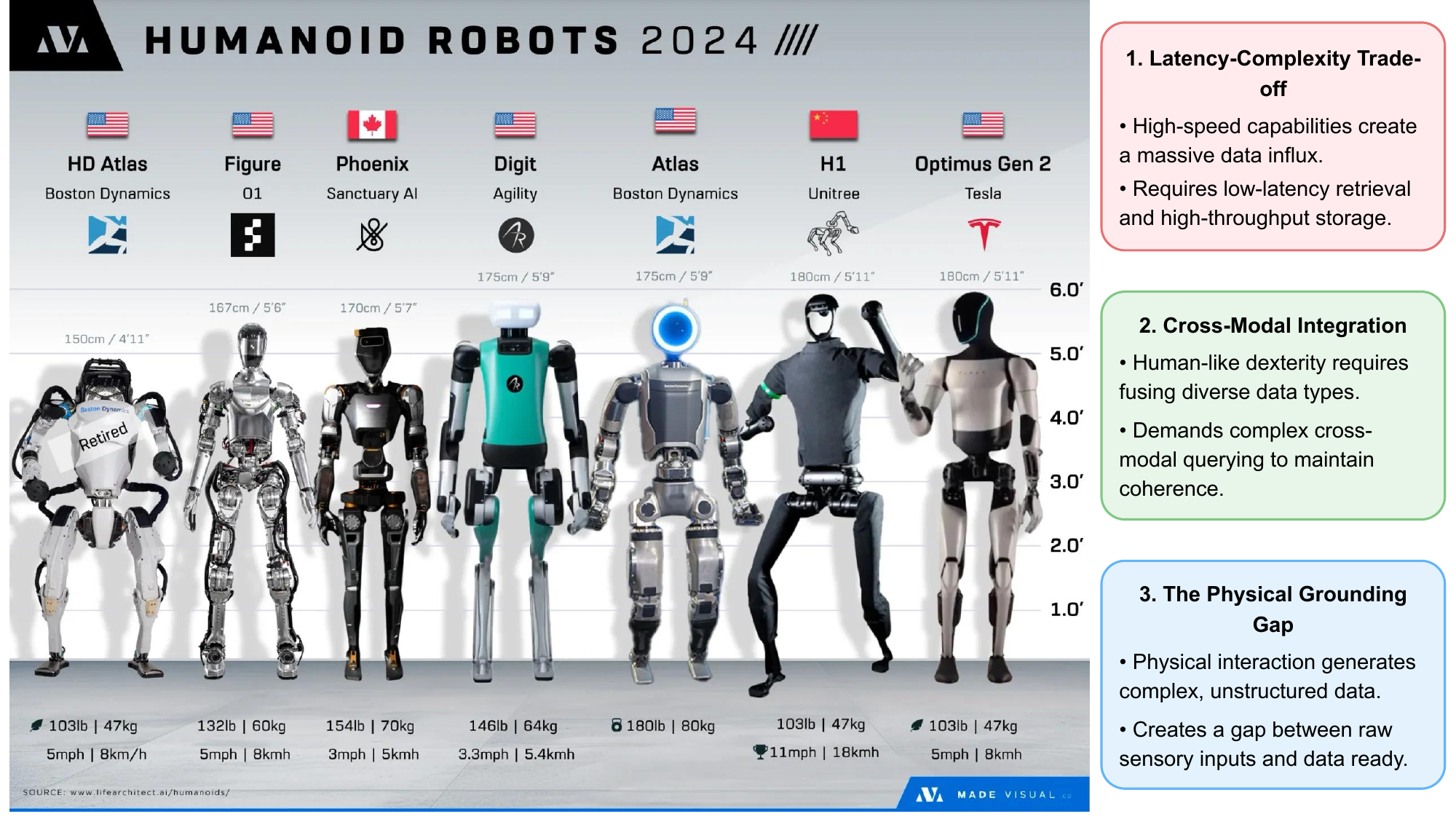}
    \caption{From robotic specifications to data management challenges. Building on an infographic of current humanoid robots (Source:~\cite{MadeVisual2024Humanoids}), this figure illustrates how key performance metrics-such as high-speed mobility and human-like dexterity-create significant challenges for multimodal data storage and retrieval, as detailed in the three panels below.}
    \label{From_robotic_specifications_to_data_management_challenges}
\end{figure*}

As EAI agents become more advanced, multimodal data plays an increasingly pivotal role. The capabilities of cutting-edge humanoid robots, as illustrated by the specifications in Fig.~\ref{From_robotic_specifications_to_data_management_challenges} , are fundamentally enabled by the real-time fusion of high-dimensional data streams like vision, proprioception, and haptic feedback. These concurrent data streams are essential for complex tasks such as balancing, navigation, and dynamic decision-making. The sheer volume, velocity, and variety of this data, however, impose significant challenges on underlying storage and retrieval systems---a core theme of this survey.

Specifically, multimodal data enhances an EAI agent's capabilities in four crucial aspects, which we detail below.

\subsubsection{Enhancing Environmental Awareness}

A primary role of multimodal data is to enhance an EAI agent's environmental awareness, which is critical for informed decision-making. Autonomous driving provides a clear illustration: visual data from cameras provides information on road signs, traffic lights, and other vehicles~\cite{wang2023surveydatasetsdecisionmakingautonomous}, while auditory signals like sirens alert the system to off-camera emergencies. Concurrently, haptic data from vibration sensors can inform the vehicle about road conditions. The fusion of these modalities creates a robust and comprehensive world model, leading to significantly improved safety and reliability compared to any single-modality system.

\subsubsection{Improving Task Comprehension}

Multimodal data is essential for an agent to accurately comprehend complex, often underspecified, human tasks. Consider a medical assistance robot: It might receive a verbal command like ``assist in repositioning the patient,'' while simultaneously interpreting a doctor's pointing gesture to understand the target location~\cite{chen2021yourefitembodiedreferenceunderstanding}. Haptic feedback ensures the applied force is safe, while visual analysis of the patient's facial expression can provide real-time feedback on their comfort level~\cite{wang2023medical}. This integration of verbal instructions, non-verbal cues, and physical feedback allows the system to disambiguate commands and execute tasks with greater precision and safety.

\subsubsection{Optimizing Decision-Making and Action}

Effective decision-making and action execution in EAI rely on a rich synthesis of perceptual data and task understanding. In industrial automation, for example, vision provides the position, shape, and color of a workpiece, guiding a robot's grasp. Haptic feedback then modulates the gripping force to prevent damage. Concurrently, the system can receive verbal instructions from a human operator, such as ``pick up a specific workpiece''~\cite{gustavsson2017human}. In hazardous environments, olfaction might even detect a chemical leak, triggering an immediate safety protocol. This fusion of multiple data streams enables more precise, efficient, and safer industrial operations.

\subsubsection{Enhancing User Interaction}
Natural and effective human-robot interaction is fundamentally multimodal. In intelligent education, for example, an agent can respond to a student's verbal questions through speech recognition~\cite{lee2011effectiveness}. Simultaneously, its vision system can assess the student's engagement level by analyzing their facial expressions and posture. Haptic interfaces allow for tactile interaction, such as a student manipulating a virtual object on a screen. By integrating these channels, the EAI system can more accurately infer the user's intent and emotional state, delivering a more personalized and adaptive experience.

In summary, multimodal data plays an irreplaceable role in embodied intelligence. By fusing data from multiple modalities, embodied intelligence systems can more accurately perceive the environment, understand tasks, make decisions, execute actions, and interact with humans. As fusion technologies advance, the performance and application scope of EAI systems will expand. However, this progress is fundamentally gated by the ability of the underlying data infrastructure to store, retrieve, and synchronize these complex data streams efficiently, a challenge we explore in the subsequent sections.

\begin{figure}
    \centering
    \includegraphics[width=1\linewidth]{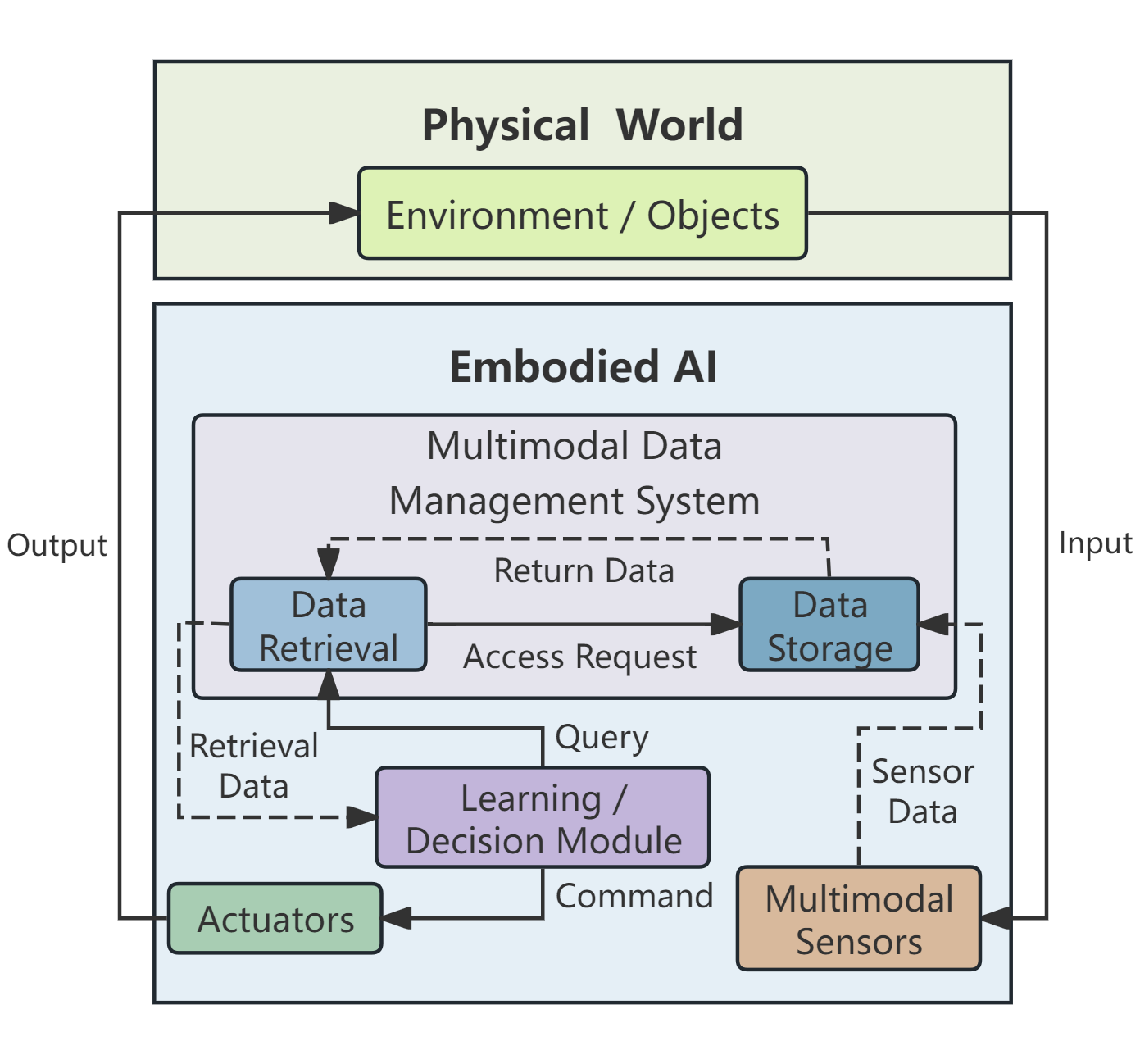}
    \caption{Conceptual Architecture and Interaction Loop of an Embodied AI System}
    \label{Conceptual_Architecture_and_Interaction_Loop_of_an_Embodied_AI_System}
\end{figure}

\section{Multimodal Data Storage}
\label{Multimodal_Data_Storage}

Embodied AI agents operate on a continuous perception-action cycle (see Fig.~\ref{Conceptual_Architecture_and_Interaction_Loop_of_an_Embodied_AI_System}), which is underpinned by a Multimodal Data Management System. This system performs two core functions: Data Storage, responsible for ingesting and organizing vast streams of environmental sensor data from the agent's interactions, and Data Retrieval, tasked with supplying relevant, timely information to the agent's learning and decision-making modules. This section focuses on the first of these pillars: the technologies and architectures for multimodal data storage in the specific context of Embodied AI.

Multimodal data storage for EAI must balance the competing demands of ultra-low-latency access for an agent's real-time control and the scalable management of massive, interconnected heterogeneous data generated over its lifespan~\cite{wang2024robotsonenewstandard}. To navigate this complex landscape, we categorize the prevailing storage paradigms into five distinct classes, each offering a unique approach to this challenge:
1) Graph Databases;
2) Multi-Model Databases;
3) Data Lakes;
4) Vector Databases;
5) Time-Series Databases.

\begin{figure*}[t!]
\centering

\resizebox{\textwidth}{!}{
\begin{minipage}{29cm} 
\centering

\begin{tikzpicture}[
    font=\sffamily,
    every node/.style={font=\large}, 
    scale=1.0
]
    \def\timelineScale{1.7} 
    \def\startyear{2010}
    \def\totalyears{16}

    \draw[-{Latex[length=4mm, width=3mm]}, line width=1.8pt, color=black!75] 
        (0,0) -- (\totalyears*\timelineScale, 0);

    \foreach \year in {\startyear,...,2025} {
        \pgfmathsetmacro{\xpos}{(\year - \startyear) * \timelineScale}
        \ifthenelse{\equal{\year}{2010} \OR \equal{\year}{2015} \OR \equal{\year}{2020} \OR \equal{\year}{2025}}{
            \draw[line width=1.5pt, color=black!80] (\xpos, 0.25) -- (\xpos, -0.25);
            \node[below=3pt, font=\large\bfseries, color=black!85] at (\xpos, -0.25) {\year};
        }{
            \draw[line width=0.8pt, color=black!60] (\xpos, 0.15) -- (\xpos, -0.15);
            \node[below=2pt, font=\normalsize, color=black!65] at (\xpos, -0.15) {'\pgfmathparse{int(mod(\year,100))}\pgfmathresult};
        }
    }

    \newcommand{\modernTimelineEvent}[5]{
        \pgfmathsetmacro{\eventx}{#1}
        \ifthenelse{\equal{#2}{up}}{
            \draw[line width=1.2pt, color=#4!80] (\eventx, 0.05) -- (\eventx, #3);
            \node[above=2pt, align=center] at (\eventx, #3) {
                {\bfseries\color{#4!90} #5}
            };
        }{
            \draw[line width=1.2pt, color=#4!80] (\eventx, -0.05) -- (\eventx, -#3);
            \node[below=2pt, align=center] at (\eventx, -#3) {
                {\bfseries\color{#4!90} #5}
            };
        }
    }

    \modernTimelineEvent{0.17 * \timelineScale}{up}{1.32}{SciPurple}{PQ~\cite{jegou2010product}}
    \modernTimelineEvent{0.5 * \timelineScale}{down}{1.54}{SciBlue}{HypergraphDB~\cite{iordanov2010hypergraphdb}}
    \modernTimelineEvent{3.4 * \timelineScale}{up}{2.3}{SciBlue}{Trinity~\cite{shao2013trinity}}
    \modernTimelineEvent{3.72 * \timelineScale}{down}{1.3}{SciPurple}{OPQ~\cite{ge2013optimized}}
    \modernTimelineEvent{4.17 * \timelineScale}{up}{1.23}{SciOrange}{Sinew~\cite{tahara2014sinew}}
    \modernTimelineEvent{4.5 * \timelineScale}{down}{2.65}{SciGreen}{Kappa Arc.~\cite{kreps2014lambda}}
    \modernTimelineEvent{4.94 * \timelineScale}{up}{2.89}{SciOrange}{NoAM~\cite{bugiotti2014database}}
    \modernTimelineEvent{5.58 * \timelineScale}{down}{1.1}{SciGray}{Gorilla~\cite{10.14778/2824032.2824078}}
    \modernTimelineEvent{6.08 * \timelineScale}{up}{1.26}{SciGray}{BTrDB~\cite{andersen2016btrdb}}
    \modernTimelineEvent{6.28 * \timelineScale}{down}{2.12}{SciGreen}{Data-Pond Arc.~\cite{inmon2016data}}
    \modernTimelineEvent{6.67 * \timelineScale}{up}{2.46}{SciBlue}{GraphJet~\cite{sharma2016graphjet}}
    \modernTimelineEvent{7.42 * \timelineScale}{down}{3.44}{SciGreen}{Functional Data Lake Arc.~\cite{jarke2017warehouses}}
    \modernTimelineEvent{8.42 * \timelineScale}{up}{1.31}{SciGray}{ModelarDB~\cite{jensen2018modelardb}}
    \modernTimelineEvent{9.15 * \timelineScale}{down}{1.2}{SciOrange}{UniBench~\cite{zhang2019unibench}}
    \modernTimelineEvent{9.5 * \timelineScale}{up}{2.55}{SciPurple}{DiskANN~\cite{jayaram2019diskann}}
    \modernTimelineEvent{9.74 * \timelineScale}{down}{2.64}{SciGreen}{Multi-Zone Functional Arc.~\cite{ravat2019data}}
    \modernTimelineEvent{10.67 * \timelineScale}{up}{1.32}{SciPurple}{ScaNN~\cite{guo2020accelerating}}
    \modernTimelineEvent{11.78 * \timelineScale}{down}{1.32}{SciBlue}{MillenniumDB~\cite{vrgoc2021millenniumdb}}
    \modernTimelineEvent{11.92 * \timelineScale}{up}{1.97}{SciPurple}{SPANN~\cite{chen2021spannhighlyefficientbillionscaleapproximate}}
    \modernTimelineEvent{13.17 * \timelineScale}{down}{2.28}{SciGreen}{Symphony~\cite{chen2023symphony}}
    \modernTimelineEvent{14.30 * \timelineScale}{up}{2.53}{SciGray}{Apache TsFile~\cite{zhao2024apache}}
    \modernTimelineEvent{14.67 * \timelineScale}{down}{1.48}{SciOrange}{MQRLD~\cite{sheng2024mqrldmultimodaldataretrieval}}
    \modernTimelineEvent{14.95 * \timelineScale}{up}{1.35}{SciOrange}{X-Stor~\cite{lei2024x}}

\end{tikzpicture}

\vspace{0.4cm} 

\newcommand{\modernlegend}[2]{
    \tikz[baseline=-0.5ex]{
        \fill[#1] (0,0) rectangle (0.25, 0.15);
    }
    \hspace{0.3em}
    \large\textbf{#2}
}

\begin{tabular}{@{}l@{\hspace{1.5cm}}l@{\hspace{1.5cm}}l@{\hspace{1.5cm}}l@{\hspace{1.5cm}}l@{}}
    \modernlegend{SciBlue}{Graph Database} &
    \modernlegend{SciPurple}{Vector Database} &
    \modernlegend{SciOrange}{Multi-model Database} &
    \modernlegend{SciGray}{Time-Series Databases} &
    \modernlegend{SciGreen}{Data Lake}
\end{tabular}

\end{minipage}
} 

\caption{Chronological Milestones in Multimodal Data Storage Technologies (2010–2024)}
\label{Chronological_Milestones_in_Multimodal_Data_Storage_Technologies}
\end{figure*}
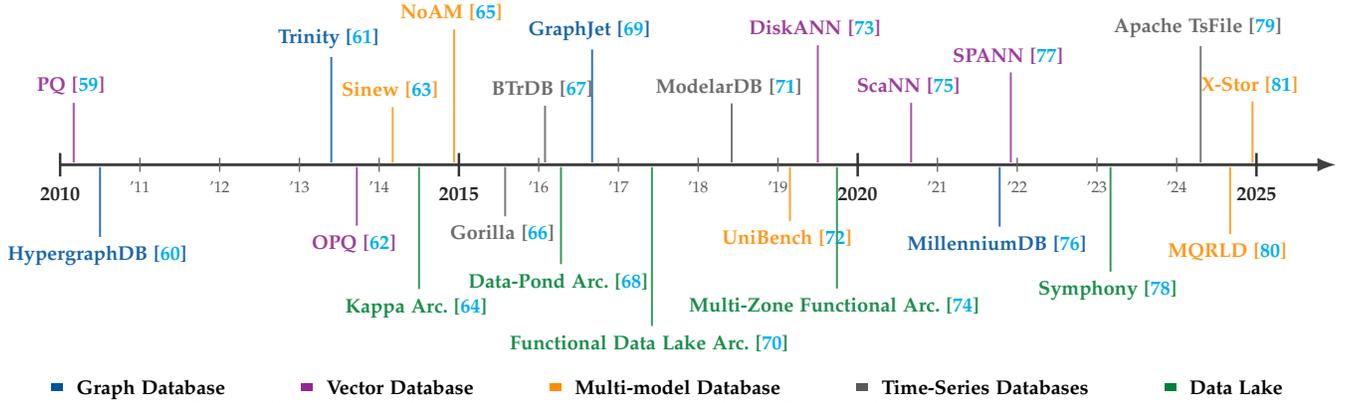

Fig.~\ref{Chronological_Milestones_in_Multimodal_Data_Storage_Technologies} provides a chronological overview of key milestones for these architectures, offering a historical context for the detailed analysis that follows, where we will examine each paradigm's suitability for EAI.

\subsection{Storage Architecture Taxonomy}  

We categorize storage solutions along two axes-data model and access pattern-into the following five paradigms:  
\begin{enumerate}[leftmargin=*]
    \item \textbf{Graph Databases:} Model complex relationships directly via nodes and edges.  
    \item \textbf{Multi-model Databases:} Natively support relational, document, graph, and other data models within a single platform. 
    \item \textbf{Data Lakes:} Employ a ``schema-on-read'' approach to retain raw data.
    \item \textbf{Vector Databases:} Optimize indexing for high-dimensional vector similarity searches.
    \item \textbf{Time-Series Databases:} Provide efficient write and temporal query capabilities tailored to voluminous sensor streams.
\end{enumerate}

The following subsections examine the core characteristics of each architecture. Table~\ref{Comparative_Summary_of_Multimodal_Storage_Architectures_for_Embodied_AI} provides a high-level comparative analysis that will serve as a guide for the detailed discussion, and a final consolidated comparison is presented in Section 3.7.

\begin{table*}[htbp]
  \centering
  \small
  
  \newcommand{\archcite}[2]{#1\par\small#2}
  
  \caption{Comparative Summary of Multimodal Storage Architectures for Embodied AI}
  \label{Comparative_Summary_of_Multimodal_Storage_Architectures_for_Embodied_AI}
  \resizebox{\textwidth}{!}{
  \begin{tabular}{
      >{\centering\arraybackslash}m{3.1cm}
      >{\centering\arraybackslash}m{2.35cm}
      >{\centering\arraybackslash}m{3.6cm}
      >{\raggedright\arraybackslash}m{4.8cm}
      >{\raggedright\arraybackslash}m{4.8cm}
      >{\raggedright\arraybackslash}m{5.0cm} 
    }
    \toprule
    \multicolumn{1}{c}{\textbf{Storage Architecture}} & 
    \multicolumn{1}{c}{\textbf{Data Model}} & 
    \multicolumn{1}{c}{\textbf{Use Case in EAI}} & 
    \multicolumn{1}{c}{\textbf{Advantages in EAI}} & 
    \multicolumn{1}{c}{\textbf{Limitations in EAI}} & 
    \multicolumn{1}{c}{\textbf{Typical Systems}} \\
    \midrule
    
    \archcite{Graph Databases}{\cite{7148480, lopez2022review}}
    & Graph & Environmental relation modeling and reasoning & 
    \begin{itemize}[leftmargin=*, topsep=0pt, itemsep=2pt, parsep=0pt, partopsep=0pt]
      \item Real-time scene correlation and analysis
      \item Supports decision-making and online learning
      \item Low-latency complex queries
    \end{itemize}
    &
    \begin{itemize}[leftmargin=*, topsep=0pt, itemsep=2pt, parsep=0pt, partopsep=0pt]
      \item Difficulty maintaining consistency under frequent, real-time updates
      \item Multimodal data mapping and association difficulties
      \item Insufficient dynamic query optimization
    \end{itemize}
    & 
    \begin{itemize}[leftmargin=*, topsep=0pt, itemsep=2pt, parsep=0pt, partopsep=0pt]
      \item HyperGraphDB~\cite{iordanov2010hypergraphdb}
      \item Trinity~\cite{shao2013trinity}
      \item GraphJet~\cite{sharma2016graphjet}
      \item MillenniumDB~\cite{vrgoc2021millenniumdb}
    \end{itemize} \\

    \archcite{Multi-model Databases}{\cite{Lu2017MultimodelDM, 10.1145/3323214}}
    & Unified multiple data models & Unified management of heterogeneous perceptual data & 
    \begin{itemize}[leftmargin=*, topsep=0pt, itemsep=2pt, parsep=0pt, partopsep=0pt]
      \item Unified management of heterogeneous data
      \item Reduces transformation and movement overhead
      \item Balances real-time and throughput
    \end{itemize}
    & 
    \begin{itemize}[leftmargin=*, topsep=0pt, itemsep=2pt, parsep=0pt, partopsep=0pt]
      \item Pressure on real-time interaction and updates
      \item High latency for complex cross-model queries
      \item Global consistency challenges
      \item Lack of dynamic context awareness
    \end{itemize}
    & 
    \begin{itemize}[leftmargin=*, topsep=0pt, itemsep=2pt, parsep=0pt, partopsep=0pt]
      \item Sinew~\cite{tahara2014sinew}
      \item NoAM~\cite{bugiotti2014database}
      \item UniBench~\cite{zhang2019unibench}
    \end{itemize} \\

    \archcite{Data Lakes}{\cite{hai2023data, computers13070183}}
    & Schema-on-read & Massive raw data storage & 
    \begin{itemize}[leftmargin=*, topsep=0pt, itemsep=2pt, parsep=0pt, partopsep=0pt]
      \item Maximizes raw data retention
      \item Supports data lineage and re-analysis
      \item Aids agent decision optimization
    \end{itemize}
    & 
    \begin{itemize}[leftmargin=*, topsep=0pt, itemsep=2pt, parsep=0pt, partopsep=0pt]
      \item Insufficient real-time performance
      \item No native cross-modal fusion
      \item Difficulty capturing dynamic context information
    \end{itemize}
    & 
    \begin{itemize}[leftmargin=*, topsep=0pt, itemsep=2pt, parsep=0pt, partopsep=0pt]
      \item Kappa Arch.~\cite{kreps2014lambda}
      \item Data-Pond Arch.~\cite{inmon2016data}
      \item Functional Data Lake Arch.~\cite{jarke2017warehouses}
      \item Multi-Zone Functional Arch.~\cite{ravat2019data}
    \end{itemize} \\

    \archcite{Vector Databases}{\cite{pan2024survey, kukreja2023vector}}
    & High-dimensional vector indexing & Multimodal semantic matching &
    \begin{itemize}[leftmargin=*, topsep=0pt, itemsep=2pt, parsep=0pt, partopsep=0pt]
      \item Semantic-level similarity search
      \item Supports incremental learning and knowledge evolution
      \item Unifies multimodal perception
    \end{itemize}
    &
    \begin{itemize}[leftmargin=*, topsep=0pt, itemsep=2pt, parsep=0pt, partopsep=0pt]
      \item High overhead for real-time updates
      \item Limited support for complex associative queries
      \item Difficulty aligning dynamic data with static indexes
    \end{itemize}
    & 
    \begin{itemize}[leftmargin=*, topsep=0pt, itemsep=2pt, parsep=0pt, partopsep=0pt]
      \item PQ~\cite{jegou2010product}
      \item OPQ~\cite{ge2013optimized}
      \item DiskANN~\cite{jayaram2019diskann}
      \item ScaNN~\cite{guo2020accelerating}
    \end{itemize} \\

    \archcite{Time-Series Databases}{\cite{Liu2023TimeSeriesSurvey}}
    & Time series & Continuous time-series sensor data processing &
    \begin{itemize}[leftmargin=*, topsep=0pt, itemsep=2pt, parsep=0pt, partopsep=0pt]
      \item High-throughput sensor stream handling
      \item Efficient time-series retrieval and analysis
      \item Reduced long-term storage cost
      \item Optimized EAI behavior strategies
    \end{itemize}
    &
    \begin{itemize}[leftmargin=*, topsep=0pt, itemsep=2pt, parsep=0pt, partopsep=0pt]
      \item Difficulty fusing heterogeneous multimodal data
      \item Real-time performance under high load not guaranteed
    \end{itemize}
    &   
    \begin{itemize}[leftmargin=*, topsep=0pt, itemsep=2pt, parsep=0pt, partopsep=0pt]
      \item Gorilla~\cite{10.14778/2824032.2824078}
      \item BTrDB~\cite{andersen2016btrdb}
      \item ModelarDB~\cite{jensen2018modelardb}
      \item Apache TsFile~\cite{zhao2024apache}
    \end{itemize} \\
    \bottomrule
  \end{tabular}
  }
\end{table*}

\subsection{Graph Databases}
\label{Graph_Databases}

Graph databases leverage graph-theoretic principles to model information, representing data entities as nodes and their intricate relationships as edges~\cite{7148480,lopez2022review}. This structure is inherently superior to traditional relational systems for managing complex, semi-structured, and highly interconnected data. Its relevance to Embodied AI is profound, as an agent's understanding of the world is not a collection of independent facts but a web of interconnected concepts. In the EAI context, graph databases provide a foundational substrate for an agent's world model, capable of intuitively encoding the rich spatial configurations of objects, the semantic meanings they hold, and the complex interaction dynamics between the agent and its environment.

To represent this complex world, three dominant graph models offer distinct approaches for balancing expressivity and utility for an EAI agent~\cite{lopez2022review}. \textbf{RDF Graphs} employ a formal Subject-Predicate-Object triple structure to bolster semantic expressivity and knowledge linkage, a rigorous framework crucial for an agent's ability to build a formal knowledge base that supports complex logical inference to uncover object affordances and deep causal relationships. \textbf{Property Graphs} provide high flexibility by allowing nodes and edges to carry arbitrary attribute key-value pairs~\cite{anuyah2024understandinggraphdatabasescomprehensive}, empowering an agent to store rich, contextual, and dynamic metadata directly on the graph by annotating an object node with its observed physical weight or a door node with its current \texttt{isOpen} state. \textbf{Hypergraphs} generalize edges into hyperedges connecting multiple nodes, thus natively modeling many-to-many relationships~\cite{iordanov2010hypergraphdb,ausiello2017directed}; this capability is vital for an EAI agent to represent complex events or group activities where a single command might trigger a coordinated action by multiple robotic arms on one workpiece.

A core advantage of graph databases, beyond their modeling capabilities, lies in their ``index-free adjacency'' query strategy. Query performance scales with the locality of the traversal rather than the global data size, a property that is critical for real-time agents~\cite{robinson2015graph,pokorny2015graph}. This empowers an EAI agent to efficiently explore its immediate context-perhaps identifying all reachable objects for a manipulation task-without being penalized by the vastness of its lifelong accumulated knowledge. This efficient pattern-matching is fundamental for real-time associative retrieval from its evolving world model.

Despite these strengths, the continuous and real-time nature of EAI exposes significant challenges for graph databases. An agent's platform demands constant, low-latency updates to mirror a perpetually shifting environment, yet existing partitioning and distributed transaction schemes often struggle to guarantee both global consistency and high performance. Moreover, the task of mapping heterogeneous multimodal sensory streams into a unified graph schema, while preserving crucial cross-modal semantics and spatiotemporal coherence, remains a major open research problem. Finally, an agent's need to dynamically optimize queries based on its runtime context calls for novel, context-aware planning strategies that go beyond the static execution models inherent to most classical graph systems.

\subsection{Multi-Model Databases}
\label{Multi-Model_Databases}

Multi-model databases natively support multiple data models, such as relational, document, and graph structures, within a single, unified system~\cite{BIMONTE2022101734,10.1145/3323214}. Their direct relevance to EAI stems from an agent's inherently multimodal and multi-structured experience of the world. Such a database offers the potential to store visual frames as documents, spatial relationships as a graph, and sensor readings as time-series data, all within one consolidated engine, thereby streamlining the complexity of the agent's data architecture~\cite{Demurjian1989TheMD}. This integrated approach contrasts sharply with solutions requiring multiple disparate database systems, significantly reducing the overhead associated with data transformation and inter-system synchronization.

According to~\cite{Lu2017MultimodelDM,10.1145/3323214}, architectural strategies for this fall into two main paradigms. \textbf{Polyglot Persistence} employs separate, specialized database systems for each data type~\cite{9245178}; while optimizing performance for individual modalities, it forces an agent's data to be scattered, introducing significant complexity for holistic tasks requiring cross-database queries and sophisticated data integration and escalating both development and maintenance costs~\cite{info10040141,kolev2016cloudmdsql,8843446}. In contrast, \textbf{Multi-Model Database Systems} embed various data models into one unified instance supporting seamless cross-model queries, a paradigm aligning well with the EAI requirement for consolidated data management by overcoming the fragmented nature of Polyglot Persistence and improving both usability and the reliability of the agent's world model~\cite{BIMONTE2022101734,mihai2020multi}.

The primary strength of an integrated multi-model database for an EAI agent is its ability to manage heterogeneous data within a single transactional context, which reduces schema conversion overheads and simplifies data pipelines. By leveraging diverse, on-demand indexing strategies for specific data models, these systems can concurrently support the low-latency responses needed for real-time perception and the high-throughput processing required for offline analysis of an agent's experiences, thus catering to multifaceted requirements like cross-modal decision-making and online learning.

Nevertheless, the demanding real-time nature of EAI scenarios exposes critical limitations. Complex cross-model joins and queries can be susceptible to unpredictable latency fluctuations, stemming from challenges in query plan scheduling and resource allocation, which is unacceptable for an agent requiring swift reactions. Furthermore, discrepancies in transaction isolation mechanisms among different data models make it difficult to implement global atomicity, potentially compromising the consistency of the agent's world state. More critically, the semantic interpretation of an agent's data is frequently contingent on dynamic context, yet current multi-model databases generally lack built-in functionalities for adaptive schema evolution or context-aware query switching, constraining their ability to process fluid, real-world semantics.

\subsection{Data Lakes}
\label{Data_Lakes}

Data lakes function as highly scalable storage platforms designed to retain vast quantities of diverse structured and unstructured data in their native, raw formats~\cite{sawadogo2021data}. This ``schema-on-read'' philosophy is particularly beneficial for EAI, as it can ingest the massive, unpredictable, and heterogeneous multimodal data streams generated from an agent's continuous interaction with the physical world without requiring predefined schemas. The core premise is to centralize all of an agent's raw experiential data to furnish a unified access point for a wide range of subsequent exploration and analytical tasks. A fully realized data lake architecture typically comprises three layers: ingestion, maintenance, and exploration, which together manage the lifecycle of this data~\cite{hai2023data}.

Various architectural paradigms for data lakes have evolved to address complex data processing needs in EAI scenarios~\cite{computers13070183}. Key evolutionary stages began with \textbf{Single-Zone Architectures}, the earliest form, which stored all data in one repository but lacked robust real-time processing and governance mechanisms, rendering them ill-suited for EAI's demanding requirements~\cite{ravat2019data}. These were followed by the \textbf{Lambda and Kappa Architectures}; the Lambda Architecture~\cite{liu2020big} introduced parallel batch and speed layers for both historical analytics and real-time updates, while the Kappa Architecture~\cite{kreps2014lambda} later simplified this by focusing solely on a streaming layer to optimize responsiveness, a crucial factor for reactive EAI agents. Further advancements led to \textbf{Partitioned Governance Models} like the Data-Pond~\cite{inmon2016data}, Zone-Based~\cite{gorelik2019enterprise,zikopoulos2015big,madsen2015build,ravat2019data}, and Functional Data Lake architectures~\cite{jarke2017warehouses}, which enhance data organization, quality monitoring, and access control through logical isolation, making them better suited for an agent's multi-stage perception and learning workflows. The most recent evolution is the \textbf{Data Lakehouse Architecture}, combining data lake flexibility with the transactional integrity and performance of data warehouses to offer a promising, more robust solution for comprehensive EAI data management.

The principal advantage of a data lake in EAI lies in its capacity for comprehensive raw data retention and administration. By persisting an agent's voluminous, multi-source outputs in their original form, it maximizes information fidelity and affords extensive reprocessing flexibility for future model training or behavior analysis. Its metadata management facilities support the efficient organization and annotation of complex interaction logs and multimodal sensory inputs, ensuring the discoverability and traceability of data throughout an agent's operational lifecycle. This enables the extraction of actionable insights from an agent's complete history to refine its behavior strategies and decision-making processes, fostering continuous performance enhancement.

However, a data lake's utility for an active EAI agent is constrained by significant challenges. The inherent latency of its typically batch-oriented architecture limits its ability to serve the real-time perception and immediate decision-making needs within an agent's physical interaction loop. While data lakes excel at accommodating heterogeneous data, their fundamental role is data retention, not real-time computation or semantic interpretation; they lack the built-in cross-modal understanding and fusion mechanisms necessary for an agent that depends on integrated multi-source perception. Additionally, the significance of an agent's data is highly contingent on dynamic context, a factor that generic data lake management paradigms cannot effectively capture or leverage, thereby limiting their value in driving context-aware embodied behaviors.

\subsection{Vector Databases}
\label{Vector_Databases}

Vector databases are specifically engineered to manage high-dimensional vector data, which can be formally a set of tuples \(v=(i, n, m)\), where \(i\) is a unique identifier, \(n\) is the vector's dimensionality, and \(m\) is a vector containing \(n\) real-valued components~\cite{kukreja2023vector}. By projecting an agent's multimodal perceptual inputs into a common high-dimensional vector space, these systems enable similarity retrieval grounded in learned semantic or geometric relationships. This approach is fundamental for EAI, as it facilitates not only the efficient indexing of vast sensory data volumes but also captures rich semantic content, underpinning subsequent tasks of multimodal fusion, inference, and decision-making.

Based on their implementation, vector database management systems (VDBMS) are classified into three categories~\cite{pan2024survey}. \textbf{Native VDBMS} like Milvus~\cite{milvus2019,wang2021milvus} and Pinecone~\cite{pinecone2021} are purpose-built for high-dimensional embeddings, furnishing an EAI agent with the real-time, low-latency queries on multimodal perceptions necessary for reactive behaviors. \textbf{Extended Systems} like pgvector~\cite{pgvector2021} and PASE~\cite{yang2020pase} augment existing database platforms with vector search capabilities, allowing an EAI deployment to consolidate semantic similarity search with other data management tasks on a single infrastructure. Lastly, \textbf{Libraries and other systems} like FAISS~\cite{faiss2017} and Annoy~\cite{annoy2014} provide high-performance Approximate Nearest Neighbor Search (ANNS) as algorithmic toolkits, offering deployment flexibility for an agent operating in resource-constrained embedded environments.

The core strength of vector databases is their ability to enable semantic-level similarity retrieval, moving beyond conventional exact-match paradigms. This empowers an EAI agent to retrieve past experiences based on conceptual similarity, allowing it to generalize from a visual query of a ``kitchen'' to retrieve memories of functionally similar but visually distinct rooms. They support streaming data ingestion and dynamic index updates, furnishing the technical bedrock for an agent's incremental learning and knowledge-base evolution. By mapping heterogeneous multimodal information into the same embedding space, they also enhance an agent's ability to perceive and interpret complex environments through cross-modal association.

Despite these capabilities, applying vector databases in EAI presents critical challenges rooted in the sim-to-real or generalization gap. Since most embedding models are trained on large, static datasets, their learned semantic spaces often fail to encompass the variability of dynamic, real-world settings an agent encounters. When an agent enters a novel environment or as its surroundings evolve, the resulting distribution shift can substantially undermine retrieval accuracy. Furthermore, the continuous motion of an embodied agent induces strong spatiotemporal correlations among its perception vectors, yet prevailing ANNS indexing algorithms largely presuppose that vectors are independent and identically distributed. Sudden perturbations in these vector distributions can therefore degrade both the performance and stability of the indexes, affecting the reliability of an agent's perception.

\subsection{Time-Series Databases}
\label{Time-Series_Databases}

Time-Series Databases (TSDBs) are a class of systems specifically engineered for the high-rate ingestion and the large-scale analysis of time-stamped data. Their fundamental strength lies in native support for the temporal dimension, enabling effective processing of data streams composed of timestamps, metrics, and associated tags~\cite{Liu2023TimeSeriesSurvey}. Unlike conventional databases that suffer from throughput bottlenecks, TSDBs employ tailored storage architectures and optimized temporal indexing to deliver specialized performance. This allows an EAI platform to achieve high-throughput storage and low-latency queries on an agent's rapidly arriving sensor data streams, empowering it to monitor its internal state and environmental dynamics in real-time to promptly adjust its behavioral policies.

The architectural paradigms for TSDBs are categorized into four principal types~\cite{Liu2023TimeSeriesSurvey}: \textbf{In-Memory TSDBs} like Gorilla~\cite{10.14778/2824032.2824078} utilize volatile storage for extremely low read/write latencies, satisfying an EAI agent's immediate control loop demands; \textbf{Relational-Backed TSDBs} such as TimescaleDB~\cite{timescale2023timescaledb} extend established databases with temporal capabilities, simplifying unified analysis of an agent's sensor inputs with its structured metadata; \textbf{Key-Value-Based TSDBs}, exemplified by OpenTSDB~\cite{opentsdb2023opentsdb}, leverage distributed key-value store features for cluster deployment, supporting massive concurrent data acquisition; and \textbf{Native TSDBs}, including Apache IoTDB~\cite{apache2023iotdb} and InfluxDB~\cite{influxdata2023influxdb}, are purpose-built for temporal data, affording efficient long-term stewardship of the vast time-series archives from an agent's lifelong learning.

The hallmark capabilities of TSDBs are high-throughput ingestion and low-latency retrieval for time-centric workloads. By employing techniques like batch writes and parallel ingestion, they can absorb the continuous sensor outputs generated by an EAI agent. Concurrently, through efficient temporal-range queries and built-in aggregation operators, they facilitate real-time retrieval and trend analysis over massive time-series repositories. For an EAI agent, this means it can swiftly access the freshest temporal insights from its recent past to refine its immediate action strategies, while also enabling long-term analysis of its behavioral patterns.

Nonetheless, integrating TSDBs into a holistic EAI ecosystem presents two principal challenges. First, the fusion of heterogeneous multimodal time-series data remains difficult. Traditional TSDBs focus on numerical streams and lack robust support for non-temporal modalities like imagery or unstructured text, hindering effective cross-modal feature association within a unified temporal framework. Second, ensuring stable performance under the stringent low-latency and high-concurrency demands of EAI is challenging. Massive sensor write loads combined with complex concurrent temporal queries can induce throughput degradation or latency variability, potentially failing to meet the hard real-time requirements of an active agent.

\subsection{Comparative Analysis of Storage Architecture Suitability}

Addressing the three fundamental requirements of EAI-namely, real-time interaction, complex relationship modeling, and massive scalability-the respective advantages and drawbacks of each storage architecture are outlined below:
\begin{itemize}[leftmargin=*]
    \item \textbf{Graph Databases:} Excel in native relationship representation, facilitating efficient subgraph queries pertinent to path planning and task reasoning. Persistent challenges include managing high-frequency sensor stream ingestion and ensuring consistency in distributed transactions.
    \item \textbf{Multi-model Databases:} Streamline heterogeneous data integration and unified querying via intrinsic multi-model support. Yet, their general-purpose query engines might exhibit suboptimal performance for specific models relative to specialized systems, potentially impacting real-time decision-making efficacy.
    \item \textbf{Data Lakes:} Enable consolidated data integration within EAI, support real-time ingestion of heterogeneous data, and possess high scalability and flexibility. However, their core design remains oriented towards large-scale batch processing, resulting in end-to-end latencies generally unsuitable for the real-time exigencies of EAI agents. Satisfying online decision-making necessitates synergistic integration with real-time stream processing frameworks or caching layers.
    \item \textbf{Vector Databases:} Facilitate multimodal semantic similarity retrieval, proving advantageous for few-shot generalization tasks. Nevertheless, the computational cost associated with real-time updates of high-dimensional indices and the extent of support for cross-model join operations warrant further enhancement.
    \item \textbf{Time-Series Databases:} Optimize the ingestion and querying of sensor stream data through mechanisms like time-slice indexing and incremental compression. They are, however, less proficient at expressing non-temporal static associations between entities, frequently necessitating amalgamation with other systems, such as graph databases, to satisfy complex query requisites.
\end{itemize}

\section{Multimodal Data Retrieval}
\label{Multimodal_Data_Retrieval}

In Embodied AI systems, data retrieval is the critical process of swiftly pinpointing relevant information from heterogeneous sources---spanning visual, linguistic, and tactile inputs---to support an agent's reasoning and action. This process must simultaneously ensure semantic comprehension and real-time responsiveness. To address these multifaceted demands, the mainstream retrieval paradigms can be classified into five principal categories:
\begin{itemize}[leftmargin=*]
    \item Fusion Strategy-Based Retrieval
    \item Representation Alignment-Based Retrieval
    \item Graph-Structure-Based Retrieval
    \item Generation Model-Based Retrieval
    \item Efficient Retrieval-Based Optimization
\end{itemize}

\begin{figure*}[t!]
\centering

\resizebox{\textwidth}{!}{
\begin{minipage}{29cm} 
\centering

\begin{tikzpicture}[
    font=\sffamily,
    every node/.style={font=\large}, 
    scale=1.0
]
    \def\timelineScale{1.7} 
    \def\startyear{2010}
    \def\totalyears{16}

    \draw[-{Latex[length=4mm, width=3mm]}, line width=1.8pt, color=black!75] 
        (0,0) -- (\totalyears*\timelineScale, 0);

    \foreach \year in {\startyear,...,2025} {
        \pgfmathsetmacro{\xpos}{(\year - \startyear) * \timelineScale}
        \ifthenelse{\equal{\year}{2010} \OR \equal{\year}{2015} \OR \equal{\year}{2020} \OR \equal{\year}{2025}}{
            \draw[line width=1.5pt, color=black!80] (\xpos, 0.25) -- (\xpos, -0.25);
            \node[below=3pt, font=\large\bfseries, color=black!85] at (\xpos, -0.25) {\year};
        }{
            \draw[line width=0.8pt, color=black!60] (\xpos, 0.15) -- (\xpos, -0.15);
            \node[below=2pt, font=\normalsize, color=black!65] at (\xpos, -0.15) {'\pgfmathparse{int(mod(\year,100))}\pgfmathresult};
        }
    }

    \newcommand{\modernTimelineEvent}[5]{
        \pgfmathsetmacro{\eventx}{#1}
        \ifthenelse{\equal{#2}{up}}{
            \draw[line width=1.2pt, color=#4!80] (\eventx, 0.05) -- (\eventx, #3);
            \node[above=2pt, align=center] at (\eventx, #3) {
                {\bfseries\color{#4!90} #5}
            };
        }{
            \draw[line width=1.2pt, color=#4!80] (\eventx, -0.05) -- (\eventx, -#3);
            \node[below=2pt, align=center] at (\eventx, -#3) {
                {\bfseries\color{#4!90} #5}
            };
        }
    }

    \modernTimelineEvent{0.17 * \timelineScale}{up}{1.37}{SciGray}{PQ~\cite{jegou2010product}}
    \modernTimelineEvent{1.67 * \timelineScale}{down}{1.78}{SciBlue}{HMF~\cite{makris2011hierarchical}}
    \modernTimelineEvent{2.55 * \timelineScale}{up}{1.98}{SciGray}{ITQ~\cite{gong2012iterative}}
    \modernTimelineEvent{3.84 * \timelineScale}{down}{1.35}{SciOrange}{DeViSE~\cite{frome2013devise}}
    \modernTimelineEvent{4.58 * \timelineScale}{up}{1.62}{SciGreen}{DeepWalk~\cite{Perozzi_2014}}
    \modernTimelineEvent{6.43 * \timelineScale}{down}{2.24}{SciGreen}{GCN~\cite{kipf2016semi}}
    \modernTimelineEvent{7.05 * \timelineScale}{up}{1.28}{SciOrange}{DGCCA~\cite{benton2017deep}}
    \modernTimelineEvent{7.20 * \timelineScale}{down}{1.23}{SciGray}{TPU~\cite{jouppi2017datacenter}}
    \modernTimelineEvent{7.75 * \timelineScale}{up}{2.78}{SciBlue}{SegNet~\cite{badrinarayanan2017segnet}}
    \modernTimelineEvent{7.88 * \timelineScale}{down}{3.16}{SciGreen}{GraphSAGE~\cite{hamilton2017inductive}}
    \modernTimelineEvent{8.80 * \timelineScale}{up}{1.75}{SciGray}{HNSW~\cite{malkov2018efficient}}
    \modernTimelineEvent{9.65 * \timelineScale}{down}{2.0}{SciBlue}{LXMERT~\cite{tan2019lxmert}}
    \modernTimelineEvent{10.58 * \timelineScale}{up}{2.50}{SciBlue}{HGMF~\cite{chen2020hgmf}}
    \modernTimelineEvent{11.16 * \timelineScale}{down}{2.76}{SciOrange}{ALIGN~\cite{jia2021scaling}}
    \modernTimelineEvent{11.38 * \timelineScale}{up}{1.31}{SciBlue}{ViLT~\cite{kim2021vilt}}
    \modernTimelineEvent{11.82 * \timelineScale}{down}{1.20}{SciOrange}{CLIP~\cite{radford2021learning}}
    \modernTimelineEvent{12.1 * \timelineScale}{up}{3.25}{SciPurple}{LaMDA~\cite{thoppilan2022lamda}}
    \modernTimelineEvent{12.5 * \timelineScale}{down}{2.34}{SciBlue}{BLIP~\cite{li2022blip}}
    \modernTimelineEvent{12.83 * \timelineScale}{up}{1.9}{SciPurple}{DSI~\cite{tay2022transformer}}
    \modernTimelineEvent{13.67 * \timelineScale}{down}{1.45}{SciPurple}{MEVI~\cite{zhang2023model}}
    \modernTimelineEvent{14.08 * \timelineScale}{up}{2.38}{SciGray}{BANG~\cite{khan2024bang}}
    \modernTimelineEvent{14.60 * \timelineScale}{down}{2.57}{SciPurple}{Embodied-RAG~\cite{xie2025embodiedraggeneralnonparametricembodied}}

\end{tikzpicture}

\vspace{0.4cm} 

\newcommand{\modernlegend}[2]{
    \tikz[baseline=-0.5ex]{
        \fill[#1] (0,0) rectangle (0.25, 0.15);
    }
    \hspace{0.3em}
    \large\textbf{#2}
}

\begin{tabular}{@{}l@{\hspace{2.5cm}}l@{}} 
    \modernlegend{SciBlue}{Fusion Strategy-Based Retrieval} &
    \modernlegend{SciPurple}{Generation Model-Based Retrieval} \\[1.2ex] 
    \modernlegend{SciOrange}{Representation Alignment-Based Retrieval} &
    \modernlegend{SciGray}{Efficient Retrieval-Based Optimization} \\[1.2ex]
    \modernlegend{SciGreen}{Graph-Structure-Based Retrieval} & 
\end{tabular}

\end{minipage}
} 

\caption{Chronological Milestones in Multimodal Data Retrieval Technologies (2010–2024)}
\label{Chronological_Milestones_in_Multimodal_Data_Retrieval_Technologies}
\end{figure*}
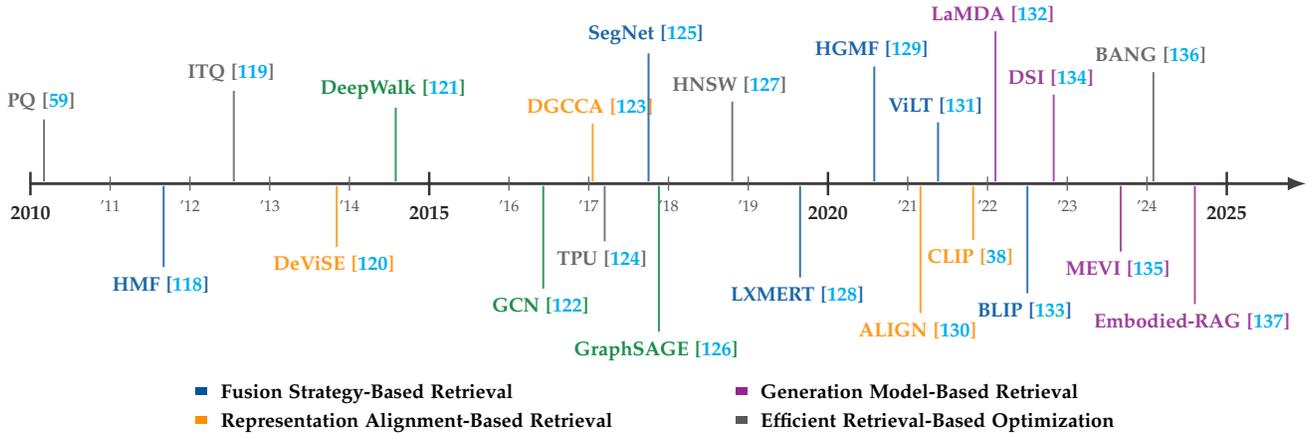

Fig.~\ref{Chronological_Milestones_in_Multimodal_Data_Retrieval_Technologies} provides a chronological overview of key milestones in these retrieval paradigms, offering a historical context for the detailed discussion that follows in the subsequent subsections.

\subsection{Retrieval Paradigm Taxonomy}

Multimodal retrieval can be organized into five core paradigms, each emphasizing a different cross-modal information acquisition mechanism:
\begin{enumerate}[leftmargin=*]
    \item \textbf{Fusion Strategy-Based Retrieval:} Integrate features from multiple modalities to exploit complementary cues, thereby enhancing retrieval accuracy and robustness.
    \item \textbf{Representation Alignment-Based Retrieval:}  Establish a unified semantic embedding space, mapping disparate modalities into that space to enable direct cross-modal similarity measurement. 
    \item \textbf{Graph-Structure-Based Retrieval:}  Explicitly model data entities and their interrelations as graph topologies, employing graph-learning techniques to capture complex dependencies for semantic retrieval. 
    \item \textbf{Generation Model-Based Retrieval:}  Leverage generative frameworks for modality conversion and data completion, improving retrieval coverage and contextual consistency. 
    \item \textbf{Efficient Retrieval-Based Optimization:}  Utilize approximate nearest-neighbor algorithms and hardware acceleration to guarantee rapid query responses over large-scale datasets. 
\end{enumerate}

The following sections analyze the core characteristics of each paradigm. Table~\ref{Comparative_Summary_of_Multimodal_Retrieval_Paradigm_for_Embodied_AI} provides a high-level comparative analysis that will serve as a guide for the detailed discussion, and a final summary is presented in Sec. 4.7.

\begin{table*}[htbp]
  \centering
  \small
  \newcommand{\archcite}[2]{#1\par\small#2}
  
  \caption{Comparative Summary of Multimodal Retrieval Paradigm for Embodied AI}
  \label{Comparative_Summary_of_Multimodal_Retrieval_Paradigm_for_Embodied_AI}
  
  \resizebox{\textwidth}{!}{
  \begin{tabular}{
      >{\centering\arraybackslash}m{3.95cm}
      >{\centering\arraybackslash}m{3.8cm}
      >{\centering\arraybackslash}m{3.8cm}
      >{\raggedright\arraybackslash}m{4.8cm}
      >{\raggedright\arraybackslash}m{4.8cm}
      >{\raggedright\arraybackslash}m{3.0cm} 
    }
    \toprule
    \multicolumn{1}{c}{\textbf{Retrieval Paradigm}} & 
    \multicolumn{1}{c}{\textbf{Core Principle}} & 
    \multicolumn{1}{c}{\textbf{Use Case in EAI}} & 
    \multicolumn{1}{c}{\textbf{Advantages in EAI}} & 
    \multicolumn{1}{c}{\textbf{Limitations in EAI}} & 
    \multicolumn{1}{c}{\textbf{Representative Algorithms}} \\
    \midrule
    
    \archcite{Fusion Strategy-Based Retrieval}{\cite{li2024multimodal, zhao2024deep, jiao2024comprehensive}}
    & Multimodal feature fusion & Multi-sensor joint perception & 
    \begin{itemize}[leftmargin=*, topsep=0pt, itemsep=2pt, parsep=0pt, partopsep=0pt]
      \item Separation of shared/private features
      \item Enhanced environmental awareness
      \item Improved cross-scenario generalization
    \end{itemize}
    &
    \begin{itemize}[leftmargin=*, topsep=0pt, itemsep=2pt, parsep=0pt, partopsep=0pt]
      \item Semantic consistency hard to guarantee
      \item Spatiotemporal alignment errors
      \item Limited real-time performance and deployment efficiency
    \end{itemize}
    & 
    \begin{itemize}[leftmargin=*, topsep=0pt, itemsep=2pt, parsep=0pt, partopsep=0pt]
      \item SegNet~\cite{badrinarayanan2017segnet}
      \item HGMF~\cite{chen2020hgmf}
      \item ViLT~\cite{kim2021vilt}
      \item BLIP~\cite{li2022blip}
    \end{itemize} \\

    \archcite{Representation Alignment-Based Retrieval}{\cite{wang2025cross, li2024multimodal}}
    & Shared semantic mapping & Precise matching of instructions and sensor data & 
    \begin{itemize}[leftmargin=*, topsep=0pt, itemsep=2pt, parsep=0pt, partopsep=0pt]
      \item Improved cross-modal consistency and robustness
      \item Enhanced holistic understanding and decision support
    \end{itemize}
    & 
    \begin{itemize}[leftmargin=*, topsep=0pt, itemsep=2pt, parsep=0pt, partopsep=0pt]
      \item Data distribution shift
      \item Spatiotemporal causal alignment difficulty
      \item Resource constraints hinder real-time
    \end{itemize}
    & 
    \begin{itemize}[leftmargin=*, topsep=0pt, itemsep=2pt, parsep=0pt, partopsep=0pt]
      \item DGCCA~\cite{benton2017deep}
      \item ALIGN~\cite{jia2021scaling}
      \item CLIP~\cite{radford2021learning}
    \end{itemize} \\

    \archcite{Graph-Structure-Based Retrieval}{\cite{peng2024graphretrievalaugmentedgenerationsurvey}}
    & Multimodal graph construction and matching & Multi-step task causal reasoning & 
    \begin{itemize}[leftmargin=*, topsep=0pt, itemsep=2pt, parsep=0pt, partopsep=0pt]
      \item Dynamic multimodal graph construction and real-time updates
      \item Efficient graph retrieval
      \item Continuous data fusion and knowledge refinement
    \end{itemize}
    & 
    \begin{itemize}[leftmargin=*, topsep=0pt, itemsep=2pt, parsep=0pt, partopsep=0pt]
      \item Limited edge-device update performance
      \item Semantic distortion in heterogeneous fusion
      \item Bias toward long-tail sparse events
    \end{itemize}
    & 
    \begin{itemize}[leftmargin=*, topsep=0pt, itemsep=2pt, parsep=0pt, partopsep=0pt]
      \item DeepWalk~\cite{Perozzi_2014}
      \item GCN~\cite{kipf2016semi}
      \item GraphSAGE~\cite{hamilton2017inductive}
    \end{itemize} \\

    \archcite{Generation Model-Based Retrieval}{\cite{li2024survey, li2024matching}}
    & Transform retrieval into generation & Filling unknown environment information & 
    \begin{itemize}[leftmargin=*, topsep=0pt, itemsep=2pt, parsep=0pt, partopsep=0pt]
      \item Generation-based completion improves robustness
      \item Context-aware adaptive retrieval
      \item RAG enhancement
    \end{itemize}
    &
    \begin{itemize}[leftmargin=*, topsep=0pt, itemsep=2pt, parsep=0pt, partopsep=0pt]
      \item Perception–language data distribution mismatch
      \item Generation latency vs. real-time response conflict
      \item Hallucination residuals; source bias risk
    \end{itemize}
    & 
    \begin{itemize}[leftmargin=*, topsep=0pt, itemsep=2pt, parsep=0pt, partopsep=0pt]
      \item LaMDA~\cite{thoppilan2022lamda}
      \item DSI~\cite{tay2022transformer}
      \item MEVI~\cite{zhang2023model}
    \end{itemize} \\
    
    \archcite{Efficient Retrieval-Based Optimization}{\cite{li2019approximate, Kachris_2025}}
    & ANN indexing + hardware acceleration & Edge-side real-time retrieval &
    \begin{itemize}[leftmargin=*, topsep=0pt, itemsep=2pt, parsep=0pt, partopsep=0pt]
      \item Efficient operation under resource and latency constraints
      \item Supports low-latency perception and real-time inference
    \end{itemize}
    &
    \begin{itemize}[leftmargin=*, topsep=0pt, itemsep=2pt, parsep=0pt, partopsep=0pt]
      \item Static resource allocation mismatches dynamic demands
      \item Index–environment misalignment
      \item Insufficient multi-timescale optimization
    \end{itemize}
    &   
    \begin{itemize}[leftmargin=*, topsep=0pt, itemsep=2pt, parsep=0pt, partopsep=0pt]
      \item PQ~\cite{jegou2010product}
      \item TPU~\cite{jouppi2017datacenter}
      \item HNSW~\cite{malkov2018efficient}
    \end{itemize} \\
    \bottomrule
  \end{tabular}
  }
\end{table*}

\subsection{Fusion Strategy-Based Retrieval}
\label{Fusion_Strategy-Based_Retrieval}

Retrieval approaches based on fusion schemes integrate multimodal descriptors to leverage the strengths of each data modality while compensating for the inherent limitations of any single modality, thereby substantially enhancing retrieval accuracy and robustness~\cite{baltruvsaitis2018multimodal,barua2023systematic,binte2024navigating}. By employing cross-modal semantic alignment and feature-space transformation techniques, these methods fully exploit the complementary information across different perceptual streams. This enables an embodied agent to construct a more precise environmental representation, such that, even in scenarios where certain sensor data are corrupted or missing, reliable perception and decision-making remain achievable through the alternative modalities.

As multimodal retrieval methodologies have evolved, traditional typologies---often categorized as early, intermediate, or late fusion---have proved insufficient, providing at best only foundational guidance~\cite{li2024multimodal,zhao2024deep,jiao2024comprehensive}. Consequently, \cite{li2024multimodal} delineates four principal fusion paradigms for modern EAI systems: \textbf{Encoder-Decoder Fusion}, a framework modularly extracting and reconstructing multimodal cues to afford EAI systems flexibility in incorporating new sensing modalities~\cite{li2018densefuse}; \textbf{Kernel-Based Fusion}, an approach utilizing kernel mappings into high-dimensional feature spaces to bolster an agent's capacity for capturing nonlinear interaction patterns in complex settings~\cite{arany2013multi,munoz2008functional}; \textbf{Graph-Based Fusion}, a method representing multimodal information as a graph structure to support an agent's long-term knowledge accumulation and transfer learning capabilities~\cite{cetin2006distributed,tong2015nonlinear,tong2017multi}; and \textbf{Attention-Driven Fusion}, a technique dynamically modulating modality weights to enable an agent to adaptively emphasize task-relevant signals while suppressing interfering noise~\cite{vaswani2017attention,yang2022continual}.

Fusion-based retrieval mechanisms effectuate a clear separation between modality-shared features and modality-specific features, reconstructing the semantic representation space of the retrieval task. The shared component facilitates cross-modal alignment by anchoring diverse perceptual inputs within a unified semantic domain; the private component preserves modality-unique information, supporting fine-grained matching. This dual-faceted strategy equips embodied agents with the capability to comprehend both the common semantics across multiple sources and the distinctive signals of each modality, thereby elevating the efficacy of cross-contextual knowledge transfer.

Nevertheless, in EAI applications, multimodal descriptors often differ in scale, statistical distribution, and semantic granularity, making it challenging to maintain semantic coherence throughout the fusion process. Moreover, sensors in EAI architectures typically operate asynchronously, resulting in timestamp misalignments that conventional fusion approaches struggle to reconcile in real time, undermining the fidelity of dynamic environment modeling. Additionally, the computational and memory demands imposed by large-scale attention mechanisms or graph-structured fusion can be prohibitive, posing significant obstacles to real-time responsiveness and edge deployment.

\subsection{Representation Alignment-Based Retrieval}
\label{Representation_Alignment-Based_Retrieval}

Representation alignment-based retrieval constructs a cross-modal shared semantic space by mapping heterogeneous data into a unified vector domain, thereby enabling the measurement of semantic similarity via spatial distance. The mathematical essence of this technique involves projecting original multimodal data into a lower-dimensional space, maximizing the spatial proximity of related data instances while minimizing the overlap among unrelated ones~\cite{wang2025cross}. This objective is often operationalized by learning mapping functions, $f_A$ and $f_B$ for two modalities, and minimizing a contrastive loss function over a batch of $N$ corresponding pairs $(x_A, x_B)$. A canonical example is the InfoNCE loss used in models like CLIP:
\begin{equation*}
L = - \sum_{i=1}^{N} \log \frac{\exp(\text{sim}(f_A(x_{A,i}), f_B(x_{B,i}))/\tau)}{\sum_{j=1}^{N} \exp(\text{sim}(f_A(x_{A,i}), f_B(x_{B,j}))/\tau)}
\end{equation*}
where $\text{sim}(\cdot, \cdot)$ is the cosine similarity and $\tau$ is a temperature parameter that controls the sharpness of the distribution. This approach effectively pulls positive pairs closer in the embedding space while pushing negative pairs apart, thus maintaining semantic consistency during cross-modal interactions and enhancing the generalization capability of embodied intelligence systems for continuous optimization of task planning and autonomous decision-making in dynamic tasks.

According to the source of supervisory signals, representation alignment-based retrieval methods are primarily categorized into two approaches. The first is \textbf{Supervised Alignment}, an approach which relies on manually annotated data for cross-modal semantic mapping. Although the annotation cost is considerable, it supports an agent in achieving precise semantic consistency for specific tasks, thereby ensuring the reliability of its decisions~\cite{ranjan2015multi,zhen2019deep}. In contrast, the second method, \textbf{Self-Supervised Alignment}, enhances system adaptability and scalability in open environments by mining latent correlations from unlabeled data, which is crucial for enabling an EAI system to possess online continual learning capabilities~\cite{feng2014cross,wang2023agree}.

This methodology aligns multimodal features via a shared embedding space, enabling unified querying and matching. Perceptual streams from each modality are processed by encoders to extract features, which are subsequently mapped to a unified vector space, thereby enhancing the consistency and robustness of cross-modal retrieval. This mechanism reinforces the comprehensive understanding capabilities of embodied intelligence systems regarding heterogeneous inputs and supports efficient decision-making responses and knowledge transfer in dynamic tasks.

Representation alignment-based retrieval methods encounter several challenges in embodied intelligence scenarios. Continuous physical interactions between the agent and the environment lead to constantly evolving distributions of perceptual data, causing performance degradation of pre-existing alignment models. Furthermore, asynchronous sampling and timestamp discrepancies among various modal sensors make it difficult for the system to capture the temporal causal relationships between actions and feedback. Moreover, as computational and storage resources on edge platforms are constrained, rendering the completion of complex representation learning and high-dimensional vector alignment under real-time constraints remains a task of considerable complexity.

\subsection{Graph-Structure-Based Retrieval}
\label{Graph-Structure-Based_Retrieval}

The retrieval paradigm grounded in graph structures employs principles from graph theory to organize and interrelate information, explicitly modeling data entities and their interactions as graph topologies and framing the retrieval task as graph traversal or subgraph matching~\cite{peng2024graphretrievalaugmentedgenerationsurvey}. By encoding multimodal perceptual inputs together with environmental entities and their interrelations into a unified graph, an embodied agent can exploit traversal-based inference to identify salient information, thereby enabling real-time situational awareness and expeditious decision-making in dynamic settings.

According to~\cite{peng2024graphretrievalaugmentedgenerationsurvey}, graph-structured retrieval techniques fall into three principal algorithmic families. \textbf{Graph Embedding Methods} aim to learn an embedding function for a given graph $G=(V,E)$, where $V$ is vertices (entities) and $E$ is edges (relationships), with the goal of learning a function $f:V \to \mathbb{R}^d$ that maps each vertex to a low-dimensional vector. Methods like DeepWalk~\cite{Perozzi_2014}, Laplacian Eigenmaps~\cite{6789755}, and GNNs~\cite{4700287} project the entire multimodal knowledge graph into a low-dimensional vector space. This process enables aligning structural knowledge with features from other modalities within a unified semantic space, furnishing a rich semantic substrate for the EAI agent~\cite{8294302}. \textbf{Graph Indexing Schemes} strike a balance between retrieval latency and semantic fidelity, supporting real-time queries over an agent's environmental knowledge base and enabling large-scale scene searches~\cite{peng2024graphretrievalaugmentedgenerationsurvey}. \textbf{Graph Matching Approaches} achieve semantic concordance between query patterns and target subgraphs by quantifying structural similarity; for an agent, task directives can be parsed into query motifs, and the environment graph searched to retrieve conforming objects~\cite{yan2016short}.

Graph-based retrieval methods transform multimodal sensory streams into an evolving graph representation whose vertices and edges are updated continuously to reflect changes in the physical world. Leveraging graph traversal and subgraph matching, an agent can anticipate the ramifications of its actions and facilitate policy transfer across disparate contexts. Moreover, newly acquired perceptual data are incrementally fused into the graph, refining the knowledge representation. Consequently, this approach not only deepens an embodied system’s comprehension of its surroundings but also undergirds long-term task planning and adaptive decision-making, bolstering robustness and flexibility in previously unseen, fluid environments.

While graph-structured retrieval offers distinct advantages for modeling multimodal entity relationships, it confronts formidable challenges within the intricate milieus of EAI. First, continual agent-environment interaction demands frequent knowledge graph updates, and reconciling this need for low-latency performance with high-precision maintenance on resource-constrained edge devices remains a primary obstacle. Second, the inherent heterogeneity of multimodal data introduces the potential for semantic distortion during cross-modal fusion, which can compromise the fidelity of the unified representation. Finally, long-tail, sparse, and unforeseen events from the real world manifest as infrequent subgraph patterns; these patterns risk being misclassified as noise, thereby inducing retrieval biases and impeding the system's generalization capabilities and resilience in open-ended environments.

\subsection{Generation Model-Based Retrieval}
\label{Generation_Model-Based_Retrieval}

The retrieval approach based on generative models reframes the search task as a text-generation problem by constructing cross-modal mappings or generating latent representations, thereby directly producing document identifiers or content to fulfill the retrieval request~\cite{li2024survey,li2024matching}. This paradigm shift steers retrieval away from pure matching toward generation, and, owing to its deep contextual understanding and cross-modal semantic representation capabilities, effectively addresses the challenges of aligning and interpreting heterogeneous perceptual data, furnishing embodied agents with real-time decision support~\cite{zhu2023large}.

According to~\cite{li2024matching}, generation model-based retrieval falls into three classes. \textbf{Document-Identifier Generation} involves approaches like DSI~\cite{tay2022transformer} and NCI~\cite{wang2022neural} that convert retrieval queries into sequence-generation tasks, a streamlined process enabling an agent to swiftly pinpoint relevant knowledge from its memory or historical data. \textbf{Content Generation} includes models like LaMDA~\cite{thoppilan2022lamda}, WebGPT~\cite{nakano2021webgpt}, and various Tool-Augmented Generation approaches~\cite{jiang2023structgpt,luo2023reasoning,schick2023toolformer,song2023restgpt}, which directly produce pertinent information to enrich system context and deepen task comprehension. Retrieval-augmented generation's principle also extends to sequential data, where models like ReMoMask~\cite{li2025remomask} generate realistic human motion, placing corresponding demands on temporal segment indexing. \textbf{Hybrid Generation} melds the strengths of the preceding two techniques: it typically employs classical retrieval to assemble a candidate set, then applies a generative model to re-rank or fuse these candidates, ultimately generating the most pertinent output. Models like MEVI~\cite{zhang2023model} and RIPOR~\cite{zeng2024scalable} exemplify this approach, balancing efficiency with precision to ensure robust information support for an EAI agent in dynamic environments.

Generative-model-based retrieval endows embodied agents with advanced cognitive and interactive faculties. By leveraging generation to complete incomplete or ambiguous inputs, the approach enhances retrieval robustness. Its context sensitivity enables systems to adapt retrieval and generation strategies flexibly in response to real-time environmental and task demands, maintaining high resilience in evolving scenarios. Moreover, through retrieval-augmented generation (RAG), embodied AI systems (e.g., ELLMER~\cite{mon2025embodied}, Embodied-RAG~\cite{xie2024embodied}) can invoke external knowledge bases during generation, effectively mitigating hallucinations and boosting output reliability, thereby reinforcing the safety and dependability of embodied AI in the mutable physical world.

Nonetheless, when applied to embodied intelligence, generative retrieval faces several challenges. Distributional discrepancies between perceptual knowledge acquired through environmental interaction and the linguistic priors learned during model pre-training impede the accurate interpretation of context-specific sensory inputs. Furthermore, the latency inherent in large-model generation conflicts with the real-time responsiveness required by such systems. Finally, akin to the well-known hallucination issue in large language models, generative retrieval methods grapple with content accuracy and trustworthiness. In addition, studies~\cite{xu2024ai,dai2023llms} indicate that retrieval systems may exhibit source bias favoring model-generated content, further jeopardizing the precision and reliability of downstream decisions.

\subsection{Efficient Retrieval-Based Optimization}
\label{Efficient_Retrieval-Based_Optimization}

The optimization driven by efficient retrieval leverages a co-design of algorithms and systems to substantially improve the query throughput and resource utilization of embodied intelligent systems within an acceptable accuracy degradation. This methodology overcomes the computational complexity and storage bottlenecks of traditional exhaustive search, endowing agents with real-time responsiveness during dynamic physical interactions and satisfying the stringent temporal requirements of on-the-fly decision making.

This methodology can be examined at two distinct levels. At the \textbf{Algorithmic Level}, approximate search techniques like LSH~\cite{gionis1999similarity}, HNSW~\cite{malkov2018efficient}, PQ~\cite{jegou2010product}, and random projection trees~\cite{dasgupta2008random} significantly reduce global search complexity, allowing an agent to find good enough information under strict time constraints~\cite{li2019approximate}. At the \textbf{Hardware Level}, technologies like FPGA-based reconfigurable computing~\cite{weisensee2004self}, GPU architectures with massive parallelism~\cite{navarro2014survey}, and near-memory computing~\cite{singh2018review} achieve remarkable gains in computational efficiency per unit of energy, providing the necessary high-throughput, low-latency hardware support for an agent's perceptual and decision-making systems~\cite{Kachris_2025}.

The efficient-retrieval optimization framework empowers embodied intelligent systems to process large-scale, multimodal datasets under tight resource and latency constraints. By trading off a controlled degree of retrieval accuracy for enhanced query speed and resource efficiency, it ensures robust operation of agents in dynamic environments. Through subspace retrieval, the computational burden is alleviated, enabling edge-side EAI devices to support low-latency perception and real-time inference, thus elevating the interaction efficiency and robustness of the agent.

Despite these advances, several challenges persist for efficient-retrieval-based optimization in EAI. Most existing frameworks assume static tasks, making it difficult to reallocate computing resources on demand to satisfy real-time responses in dynamic settings. Furthermore, once an agent alters the environment’s state, precomputed indices often become misaligned with the current scenario, necessitating environment-aware and predictive index updating mechanisms. Finally, EAI applications span from millisecond-scale reactions to hour-scale planning, yet current methods predominantly optimize for a single time horizon, hindering the balance between efficiency and accuracy across multiple temporal scales and thus impairing sustained reasoning and decision making in complex scenarios.

\subsection{Comparative Analysis of Retrieval Paradigm Suitability}

Considering the three core requirements of EAI-real-time interaction, semantic consistency, and comprehension of complex relationships-the suitability of the five paradigms is as follows:
\begin{itemize}[leftmargin=*]
    \item \textbf{Fusion Strategy-Based Retrieval:} Enhances robustness and semantic coherence by amalgamating diverse modalities, yet its computational overhead limits continuous online updates in resource-constrained EAI contexts.
    \item \textbf{Representation Alignment-Based Retrieval:} Simplifies cross-modal similarity measurement via a shared embedding space, supporting consistency across heterogeneous inputs; however, they are vulnerable to semantic drift in dynamic environments.
    \item \textbf{Graph-Structure-Based Retrieval:} Models intricate inter-entity relations to facilitate complex reasoning tasks, but graph maintenance and traversal become computationally prohibitive as scale increases. 
    \item \textbf{Generation Model-Based Retrieval:} Boosts adaptability and retrieval coverage through data generation, though inherent uncertainty may introduce noise or ``hallucinations'', posing risks to reliability and safety.
    \item \textbf{Efficient Retrieval-Based Optimization:} Achieves rapid response by trading off minor accuracy degradation, but this strategy may compromise decision-making dependability, system interpretability, and output consistency in multifaceted tasks.
\end{itemize}

\section{Discussion}
\label{Discussion}

\subsection{Key Challenges in Embodied AI Data Management}

\begin{figure*}
    \centering
    \includegraphics[width=1\linewidth]{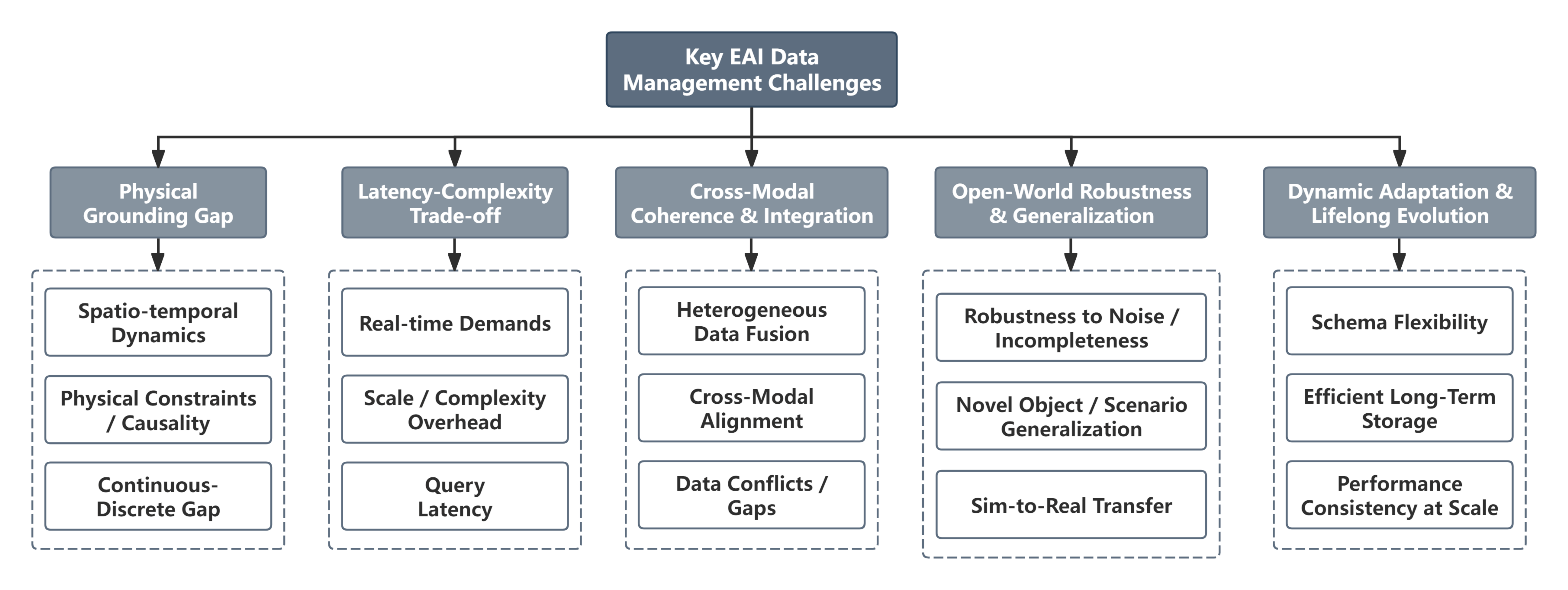}
    \caption{Taxonomy of Key Challenges for Multimodal Data Management in Embodied AI}
    \label{Taxonomy_of_Key_Challenges_for_Multimodal_Data_Management_in_Embodied_AI}
\end{figure*}

Applying established multimodal data storage and retrieval technologies to Embodied AI (EAI) presents significant hurdles, primarily because EAI systems operate through continuous physical interaction, unlike traditional AI focused on disembodied data. This discussion synthesizes the core challenges identified throughout our survey. These challenges, such as the Physical Grounding Gap, the Latency-Complexity Trade-off, and Cross-Modal Coherence \& Integration, were first introduced by the practical demands of modern humanoid robots (as illustrated in Fig.~\ref{From_robotic_specifications_to_data_management_challenges}). We now taxonomize them into five key areas, as illustrated in Fig.~\ref{Taxonomy_of_Key_Challenges_for_Multimodal_Data_Management_in_Embodied_AI}: the Physical Grounding Gap, the Latency-Complexity Trade-off, Cross-Modal Coherence \& Integration, Open-World Robustness \& Generalization, and Dynamic Adaptation \& Lifelong Evolution. In the following, we elaborate on each of these challenges.

\textbf{Representing Physical Reality in Data Structures:} A fundamental disconnect exists between the continuous, physically grounded nature of EAI experiences and the often discrete, abstract nature of data management systems. Capturing the rich spatio-temporal dynamics, inherent physical constraints, and causal links between actions and environmental changes within current database models or retrieval indices remains a major challenge. Existing storage architectures may struggle to efficiently encode this deeply contextual, embodied information, potentially losing vital details needed for robust agent behavior. Notably, this physical-grounding gap serves as the foundational bottleneck for the subsequent challenges. Lacking a unified representation of spatio-temporal dynamics directly exacerbates the latency-complexity trade-off: when models attempt high-fidelity encoding of embodied interactions, computational overhead soars, impeding real-time retrieval. Meanwhile, without a common physical reference frame, achieving cross-modal coherence becomes significantly harder---visual, haptic, and proprioceptive streams cannot be semantically aligned if they originate from incompatible abstractions.

\textbf{Latency-Complexity Trade-off in Interactive Systems:} EAI's perception-action loop demands extremely low latency for both real-time data ingestion/updates and subsequent information retrieval to enable fluid interaction and rapid decision-making. However, the multimodal data generated over an agent's lifetime is vast and complex. Scalable storage solutions and sophisticated retrieval algorithms needed to manage this data inherently introduce computational overhead and potential delays. Balancing the need for instantaneous response with the processing demands of large-scale, high-dimensional multimodal data query and analysis is a critical design tension across nearly all storage and retrieval approaches surveyed.

\textbf{Achieving Cross-Modal Coherence and Integration:} EAI agents rely on integrating information from diverse, potentially noisy, and asynchronous sensors. Ensuring that data from different modalities (vision, touch, audio, etc.) can be meaningfully related and contribute to a coherent understanding of the environment is paramount. While specific storage structures aim to accommodate heterogeneity and retrieval methods strive for semantic alignment, effectively unifying these disparate streams---preserving crucial modality-specific details while establishing robust cross-modal semantic links---remains difficult. Handling conflicting or incomplete sensor data during storage or retrieval adds another layer of complexity.

\textbf{Supporting Dynamic Adaptation and Lifelong Evolution:} EAI systems must adapt to changing environments and continuously learn from experience. This requires data management systems capable of dynamic evolution. Storage schemas need flexibility to accommodate new knowledge or changing data patterns, while retrieval models must support incremental learning without catastrophic forgetting. Furthermore, the ever-growing volume of interaction data necessitates efficient long-term storage strategies and mechanisms to maintain consistent performance as the knowledge base expands throughout the agent's lifespan.

\textbf{Ensuring Generalization and Robustness Beyond Controlled Settings:} Real-world deployment exposes EAI systems to unpredictable situations and `long-tail' events not typically encountered in training environments. Data storage must be resilient to noisy or incomplete inputs. Retrieval mechanisms, especially those based on learned representations or pattern matching, must generalize semantic understanding effectively to novel objects and scenarios. Bridging the simulation-to-reality gap and ensuring reliable data interpretation and retrieval in unstructured, open-world environments is essential for practical EAI applications.

Crucially, this challenge of generalization and robustness is intrinsically linked to the agent's capacity for dynamic adaptation and lifelong evolution. An agent cannot truly be robust in an open world without the ability to continuously update its internal models from novel experiences. This creates a symbiotic cycle: the lifelong learning mechanisms enable the agent to assimilate long-tail events, while its encounters in the open world provide the essential, unpredictable data needed to drive meaningful adaptation. Therefore, solving these two final challenges in tandem is paramount for creating agents that can move beyond controlled settings and operate autonomously in reality.

\subsection{Key Insights and Synthesis}

Our analysis shows these challenges are interdependent at the system level. These core challenges are: the Physical Grounding Gap, the Latency-Complexity Trade-off, Cross-Modal Coherence \& Integration, Dynamic Adaptation \& Lifelong Evolution, and Open-World Robustness \& Generalization. For instance, physics-agnostic data models not only undermine retrieval efficiency but also exacerbate cross-modal alignment issues; optimizing for low-latency retrieval without a holistic view can lead to resource contention in complex environments. We contend that future research must start by establishing physics-aware foundational frameworks that unify storage and retrieval designs, aiming for efficient, robust, and interpretable Embodied AI systems in real-world deployments.

\section{Conclusion}
\label{Conclusion}

This paper has surveyed the critical landscape of multimodal data storage and retrieval technologies within the context of Embodied AI (EAI). Recognizing that EAI agents learn and operate through continuous physical interaction, generating complex and heterogeneous data, we examined the suitability and inherent challenges of applying established data management techniques to this unique domain.

We reviewed diverse storage architectures, from graph and multi-model databases designed for complex relationships and heterogeneity, to data lakes, vector databases, and time-series databases optimized for scale, semantic similarity, and temporal sequences, respectively. We also investigated various retrieval paradigms, including Fusion Strategy-Based methods, Representation Alignment-Based methods, Graph-Structure-Based methods, Generation Model-Based methods, and Efficient Retrieval-Based Optimization methods. Our analysis consistently highlighted a critical tension inherent to EAI: the challenge of managing vast, complex, dynamic, and physically grounded multimodal data while satisfying stringent requirements for low-latency processing, semantic coherence, robust adaptation, and efficient resource utilization.

The five core challenges we have identified underscore that simply migrating existing technologies is insufficient. These challenges are: the Physical Grounding Gap, the Latency-Complexity Trade-off, Cross-Modal Coherence \& Integration, Open-World Robustness \& Generalization, and Dynamic Adaptation \& Lifelong Evolution. Effective data management for EAI therefore demands novel solutions and adaptations specifically tailored to its embodied nature.

Looking forward, several key research directions emerge as priorities. These directions are not isolated; rather, they form an interconnected roadmap where progress in one area often enables breakthroughs in others.

\begin{enumerate}[leftmargin=*]
    \item \textbf{Physically-Grounded Data Models and Architectures:} A foundational challenge lies in the disconnect between current data management paradigms, which are rooted in abstract feature vectors and static graphs, and the continuous, causal dynamics of the physical world. This gap prevents systems from adaptively understanding environmental changes and physical constraints in real scenarios. Most research has focused on building representations in high-dimensional embedding spaces, often overlooking the integration of object permanence, motion laws, and environmental feedback at the storage layer. Although progress has been made in simulation, real-world deployments reveal difficulties in transferability and a lack of interpretability. We therefore advocate for defining high-level, physics-aware data standards and constructing a universal framework that natively supports spatio-temporal and causal information, enabling seamless integration with real-world interaction data.
    \item \textbf{Real-time, Adaptive Storage and Retrieval Co-design:} A critical gap exists in the ability of current systems to dynamically balance latency and accuracy, as they typically rely on static, offline-defined policies. As tasks and hardware resources evolve, these systems lack a self-organizing policy to achieve a sustainable equilibrium, tending instead to overemphasize either speed or precision. Current approaches often fix indexing or caching policies during the offline phase, allowing only simple threshold-based adjustments at runtime. This approach lacks holistic resource reallocation driven by global performance metrics and task priorities, leading to unstable behavior in complex environments. Establishing unified co-design principles and performance metrics is therefore imperative to guide future systems toward a dynamic, global balancing of speed and precision, creating a scalable blueprint for adaptive resource management.
    \item \textbf{Robust Cross-Modal Representation and Alignment for Interaction:} There is a notable absence of a clear framework for the unified modeling of multimodal data under the asynchronous, noisy, and partially missing conditions typical of EAI. This deficiency hinders systems’ ability to perform cross-sensor collaborative perception and make robust decisions. While early and late fusion strategies yield results in simple scenarios, they fail to address modality mismatches and semantic drift during complex interactions. Research often stops at local alignment without high-level guidance for overall semantic consistency. There is an urgent need for a global alignment architecture, defining unified multimodal representation standards and evaluation methodologies, to build cross-scenario and cross-task generalizable alignment solutions.
    \item \textbf{Scalable Lifelong Data Management for EAI:} Lifelong EAI systems face a critical trade-off between retaining historical knowledge and managing ever-growing storage costs, a challenge that current simplistic strategies fail to address. Existing approaches typically resort to full archiving or periodic cleanup, lacking fine-grained decisions based on knowledge importance and semantic value. Consequently, they cannot effectively balance long-term knowledge accumulation with real-time access demands. We propose the development of a high-level lifelong memory management framework, which would define meta-policy interfaces and evaluation criteria to enable autonomous decisions on the forgetting, abstraction, and retention of knowledge.
    \item \textbf{Benchmarking and Standardization for EAI Data:} The lack of unified evaluation protocols and data interfaces severely impedes cumulative research progress in EAI data management. The current landscape is highly fragmented, with public datasets and toolchains that are not interoperable, and studies typically rely on custom pipelines for testing. This absence of community-endorsed benchmarks and mechanisms for continuous updates makes it difficult to establish a durable research baseline. It is therefore necessary to drive the development of open evaluation protocols and unified interface standards, building an extensible and maintainable comprehensive benchmarking platform to foster community-driven progress.
    \item \textbf{Ethical Frameworks for Embodied Data Handling:} Embodied data from human-centric environments raises significant ethical and privacy concerns that existing general-purpose models do not adequately address. Current privacy protections are ill-equipped for the diversity and continuity of online interaction data and often overlook the re-identification risks posed by unique behavioral patterns. We call for the establishment of a system-level, privacy-by-design governance framework, aligning technical standards with social responsibility guidelines to ensure the comprehensive protection of personal rights across the entire data lifecycle, from collection to storage and usage.
\end{enumerate}

Addressing these challenges through focused research will be crucial for unlocking the full potential of Embodied AI, enabling the development of more intelligent, adaptable, and reliable systems capable of operating effectively in the complexities of the real world. Ultimately, this review lays the foundation for future breakthroughs by arguing that data management must be elevated from a downstream engineering consideration to a central topic of scientific inquiry. It not only maps the current technological landscape but also pinpoints the critical scientific bottlenecks that must be overcome to realize truly autonomous, physically grounded embodied agents.

\small
\bibliographystyle{IEEEtran}
\bibliography{ref}

\end{document}